\documentclass[]{fairmeta} %

\usepackage{amsmath,amsfonts,bm}

\def\Secref#1{Section~\ref{#1}}

\def\eqref#1{equation~\ref{#1}}

\def\1{\bm{1}}

\DeclareMathAlphabet{\mathsfit}{\encodingdefault}{\sfdefault}{m}{sl}
\SetMathAlphabet{\mathsfit}{bold}{\encodingdefault}{\sfdefault}{bx}{n}

\usepackage[utf8]{inputenc}  %
\usepackage[T1]{fontenc}     %
\usepackage{hyperref}        %
\usepackage{url}             %
\usepackage{booktabs}        %
\usepackage{amsfonts}        %
\usepackage{nicefrac}        %
\usepackage{microtype}       %
\usepackage{graphicx}
\usepackage{multicol}
\usepackage{multirow}
\usepackage{xspace}
\usepackage{wrapfig}
\usepackage{enumitem}
\usepackage[dvipsnames,table]{xcolor}
\usepackage{colortbl}
\usepackage{caption}
\usepackage{lmodern} 
\usepackage{algorithm}
\usepackage{algorithmic}
\usepackage{subcaption}
\usepackage{wrapfig}
\usepackage{xspace}

\usepackage{multicol}
\usepackage{multirow}
\usepackage{pifont}
\usepackage{tabularx}
\newcolumntype{R}{>{\raggedleft\arraybackslash}X}
\definecolor{lightblue}{RGB}{220,235,250}
\definecolor{lightgray}{gray}{0.91}

\definecolor{reallab-red}{RGB}{248,191,199}
\definecolor{reallab-blue}{RGB}{59,161,255}
\definecolor{reallab-green}{RGB}{157,207,127}
\definecolor{reallab-yellow}{RGB}{249,180,13}
\definecolor{oursyellow}{RGB}{242,217,108}
\definecolor{rubrichubteal}{RGB}{168,213,213}
\definecolor{baseblue}{RGB}{43,63,123}
\newcommand{\dtp}[1]{{\footnotesize\textcolor{green!50!black}{#1}}} %
\newcommand{\dtn}[1]{{\footnotesize\textcolor{red!50!black}{#1}}}   %

\usepackage[most,skins,theorems]{tcolorbox}
\usepackage{listings}
\usepackage{fvextra}
\newtcolorbox{promptbox}[1][]{
  enhanced, breakable,
  colback=gray!1,      
  colframe=gray!60,    
  coltitle=black,      
  boxrule=2pt,
  arc=10pt,
  left=6pt, right=6pt, top=6pt, bottom=6pt,
  title={#1}, fonttitle=\bfseries,
  attach boxed title to top left={yshift*=-3mm},
  boxed title style={colback=gray!10}
}

\tcbset{
  aibox/.style={
    width=\linewidth,
    top=8pt,
    bottom=4pt,
    colback=reallab-red!15,
    colframe=reallab-red,
    colbacktitle=reallab-red!90!black,
    enhanced,
    center,
    attach boxed title to top left={yshift=-0.1in,xshift=0.15in},
    boxed title style={boxrule=0pt,colframe=white,},
  }
}
\newtcolorbox{AIbox}[2][]{aibox,title=#2,#1}

\usepackage{amsthm}
\theoremstyle{plain}

\theoremstyle{definition}

\theoremstyle{remark}

\newcommand{\methodname}{PaperGym\xspace}

\title{PaperGym: Rubric-Centered Evolution for Research-Plan Generation}

\author[1 *]{Yuhan Wang}

\author[1 *]{Zhengxi Lu}

\author[1]{Yuchen Yan}

\author[2]{Kaitao Song}

\author[1]{Wenqi Zhang}

\author[1]{Weiming Lu}

\author[1]{Jun Xiao}

\author[1]{Yueting Zhuang}

\author[1 \dagger]{Yongliang Shen}

\affiliation[1]{Zhejiang University}
\affiliation[2]{Apple}

\contribution[*]{Equal contributions}
\contribution[\dagger]{Corresponding author}

\abstract{
Research planning is the decisive capability of AI scientists. Yet a research plan admits no verifiable answer, so reinforcement learning lacks the environment it requires: tasks paired with a critic. Rubrics extracted from scientific papers can supply the critic. Existing pipelines, however, draw the question and the criteria from the same content, so the reward can be earned by paraphrase. The rubric is further compressed into a single scalar per rollout. We introduce \textbf{\methodname{}}, a unified framework that turns each research paper into a complete training environment. \methodname{} exploits the structure of a paper: the question is synthesized from the research goal and background, while the criteria are derived from the method and experiments. The criteria span methodological innovation and experimental design, and criterion leakage falls to 3.7\%, versus 11.90\% to 34.10\% in existing datasets. Training uses the rubric twice: first as privileged context for OPSD's self-teacher, then as the reward for GRPO. Across Qwen3-1.7B/4B/8B, this schedule outperforms supervised fine-tuning, either stage alone, and the reverse ordering, improving five-benchmark averages by +5.6, +5.0, and +4.8 points. With the recipe held fixed, models trained on \textbf{\methodname{}-20k} win 58.1\% of three-way comparisons, against 28.2\% for RubricHub Science. The trained Qwen3-8B reaches 73.48 on ResearchQA, above the far larger Kimi K2.6. We release the pipeline, the 20,000-instance corpus \methodname{}-20k, and the benchmarks \textbf{\methodname{}-Innov} and \textbf{\methodname{}-Design}.
}

\date{\today}
\metadata[Project Page]{\url{https://zju-real.github.io/PaperGym}}
\metadata[Code]{\url{https://github.com/ZJU-REAL/PaperGym}}
\correspondence{\email{\{zjuwangyuhan,zhengxilu,syl\}@zju.edu.cn}}

\begin{document}

\maketitle

\section{Introduction}

AI scientists have made rapid progress toward automating scientific discovery, producing complete research papers and new findings in the natural sciences~\citep{lu2024ai, mitchener2025kosmos, weng2025deepscientist}. Every such result begins with a research plan, which specifies the hypotheses, methods, and experiments to be carried out; the quality of the plan bounds what the system can ultimately achieve~\citep{goel2025training}. Unlike a mathematical answer or a program, however, a research plan admits no automatic check: there is no ground truth to compare against, and judging its quality requires expert review.

Reinforcement learning with verifiable rewards has driven remarkable progress in mathematics and code generation, where every problem is paired with an automatically checkable answer~\citep{guo2025deepseekr1, yu2026dapo}. The paradigm relies on a training environment: a collection of tasks, each accompanied by a critic that scores candidate solutions. Research planning offers no such environment. Expert review cannot provide feedback at the scale that training requires. The scientific literature, however, already contains both components at scale: millions of research papers record realistic problems together with the solutions their authors developed. This knowledge is embedded in unstructured text and cannot directly serve as training tasks or reward signals.

Every paper poses a research problem, develops a solution, and reports experiments that support it, so a realistic task can be drawn from the problem and judging criteria from the solution. Recent studies follow this route, converting papers into question-answer pairs for supervised fine-tuning~\citep{researchgpt2025} or extracting research goals with grading rubrics that serve as rewards for reinforcement learning~\citep{goel2025training, fan2026deepinnovator, sauter2026evoideator}. The environments obtained in this way, however, suffer from two problems: the reward does not faithfully measure the ability to plan, and the information it carries is largely wasted during training.

Existing pipelines draw the question and the grading criteria from a paper without regard to its structure, often from the same content~\citep{goel2025training, yifei2025researchqa}. As a result, the criteria frequently restate what the question already says: we find that 11.90\% to 34.10\% of them can be inferred from the question alone (Table~\ref{tab:leakage}). A model can thus raise its reward simply by paraphrasing the question. Moreover, a plan is judged on both the method it proposes and the experiments that validate it. Instance-specific criteria in existing datasets capture almost only the method, while experimental design is checked by generic guidelines shared across all instances~\citep{goel2025training, sauter2026evoideator}.

Even when the criteria themselves are sound, current training exploits only a small part of the supervision they provide. In rubric-as-reward training, each criterion is checked by a separate judge call, yet the verdicts reach the policy only as one aggregated scalar per rollout~\citep{gunjal2025rubrics, goel2025training}. Supervised fine-tuning sidesteps the rubric and imitates the single reference plan; because a research question admits many valid plans, this reduces output diversity and yields little gain in practice~\citep{goel2025training}. Recent work conditions a self-distillation teacher on the rubric to recover token-level guidance~\citep{gu2026rethinking, rezaei2026rubric}, but distillation alone never checks the generated plans against the criteria, and we find that it overfits the privileged context (Figure~\ref{fig:scale_and_train}).

In this paper, we introduce \textbf{\methodname{}}, a unified framework that turns each research paper into a complete training environment. The pipeline exploits the inherent structure of a paper to separate the sources of the two components: the question is synthesized only from the research goal and background, while the reference answer is derived only from the method and experimental design. This single constraint cuts criterion leakage to 3.7\%, three to nine times lower than in existing datasets. Each instance carries ten atomic binary criteria that examine both methodological innovation and experimental design. Training then extracts the full value of the rubric by using it twice in a rubric-centered evolution of the plan distribution: as privileged context for a self-distillation teacher, which converts the criteria into dense token-level guidance and builds a broad prior over valid plans, and as the reward for GRPO, which verifies complete plans against the criteria and refines the policy. Applied to 20,000 papers across computer science, physics, and economics, the pipeline yields \methodname{}-20k and two held-out benchmarks, \methodname{}-Innov and \methodname{}-Design.

Extensive experiments support both sides of the framework. With the training recipe held fixed, model trained on \methodname{}-20k win 58.1\% of three-way comparisons, against 28.2\% for the same model trained on RubricHub Science: the margin comes from the data. The two-stage schedule outperforms supervised fine-tuning, either stage alone, and the reverse ordering at all three scales, improving the five-benchmark averages of Qwen3-1.7B, 4B, and 8B by +5.6, +5.0, and +4.8 points. The trained Qwen3-8B reaches 73.48 on ResearchQA, above the far larger Kimi K2.6 at 73.19.

Our main contributions are summarized as follows:

\begin{itemize}
    \item We propose a pipeline that turns research papers into training environments, constructing tasks independently of the dual-dimension grading criteria; the resulting \methodname{}-20k contains 20,000 instances across three domains, with 3.7\% criterion leakage versus 11.90\% to 34.10\% for existing datasets.
    \item We construct \methodname{}-Innov and \methodname{}-Design, held-out benchmarks that separately evaluate the methodological innovation and the experimental design of generated research plans.
    \item We develop a two-stage schedule that uses the same rubric first as privileged context for self-distillation and then as the GRPO reward; it outperforms supervised fine-tuning, either stage alone, and the reverse ordering, and Qwen3-8B reaches 73.48 on ResearchQA, above Kimi K2.6.
\end{itemize}

\clearpage %
\section{Related Work}

\subsection{LLMs for Scientific Research}

Efforts to automate scientific research include \emph{agent-based} pipelines that coordinate dedicated modules for each research stage~\citep{lu2024ai, schmidgall2025agent} and \emph{search-based} methods that iteratively refine candidates via evolutionary or Bayesian operators~\citep{novikov2025alphaevolve, weng2025deepscientist}. Both keep the model frozen, bounding output quality by the base model's fixed capacity. A third paradigm, \emph{training-driven augmentation}, internalizes research capabilities into model parameters~\citep{he2025pasa, zeng2025reviewrl, lu2026skill0}. For research-plan generation, \citet{goel2025training} pioneered rubric-driven RL; DEEPINNOVATOR~\citep{fan2026deepinnovator} and EvoIdeator~\citep{sauter2026evoideator} follow with variant reward designs. All three, however, rely on sequence-level supervision. Our work addresses this limitation by combining rubric-based OPSD with GRPO for dense token-level guidance.

\subsection{LLM Post-Training}

Reinforcement learning from verifiable rewards (RLVR) has become the dominant post-training paradigm for reasoning~\citep{shao2024deepseekmath, yu2026dapo}. GRPO~\citep{shao2024deepseekmath} estimates advantages from group-level outcome rewards without a learned critic, yet its sequence-level scalar treats all tokens uniformly. On-Policy Self-Distillation (OPSD)~\citep{zhao2026self} provides a complementary, denser signal: the same model serves as both teacher (conditioned on the answer) and student (seeing only the problem), minimizing per-token KL divergence along on-policy rollouts. Recent work combines the two, using the teacher--student gap as a detached auxiliary loss~\citep{lu2026sdar, li2026unifying, han2026rstg} or as reward-densifying supervision~\citep{yang2026g-opd}. Our method adopts this combination, applying rubric-conditioned OPSD followed by rubric-as-rewards GRPO for both dense distributional guidance and outcome-driven optimization.

\subsection{Rubric as Rewards}

The Rubric-as-Rewards (RaR) paradigm~\citep{gunjal2025rubrics} decomposes response quality into prompt-specific criteria, offering more interpretable supervision than holistic scalar rewards. RaR faces two bottlenecks: the LLM-as-a-judge loop requires multiple calls per rollout, and the policy can exploit verifier blind spots via reward hacking~\citep{mahmoud2026reward}. Recent work converts sparse rubric rewards into dense token-level supervision via OPSD~\citep{gu2026rethinking, rezaei2026rubric}, but the rubric-conditioned teacher introduces privileged information leakage that typically requires masking or gating mechanisms. Our two-stage pipeline mitigates both issues: confining rubric evaluation to the GRPO stage reduces verifier calls, while careful separation of privileged information avoids leakage.

\section{Method}

\begin{figure*}[t]
    \centering
    \includegraphics[width=\textwidth]{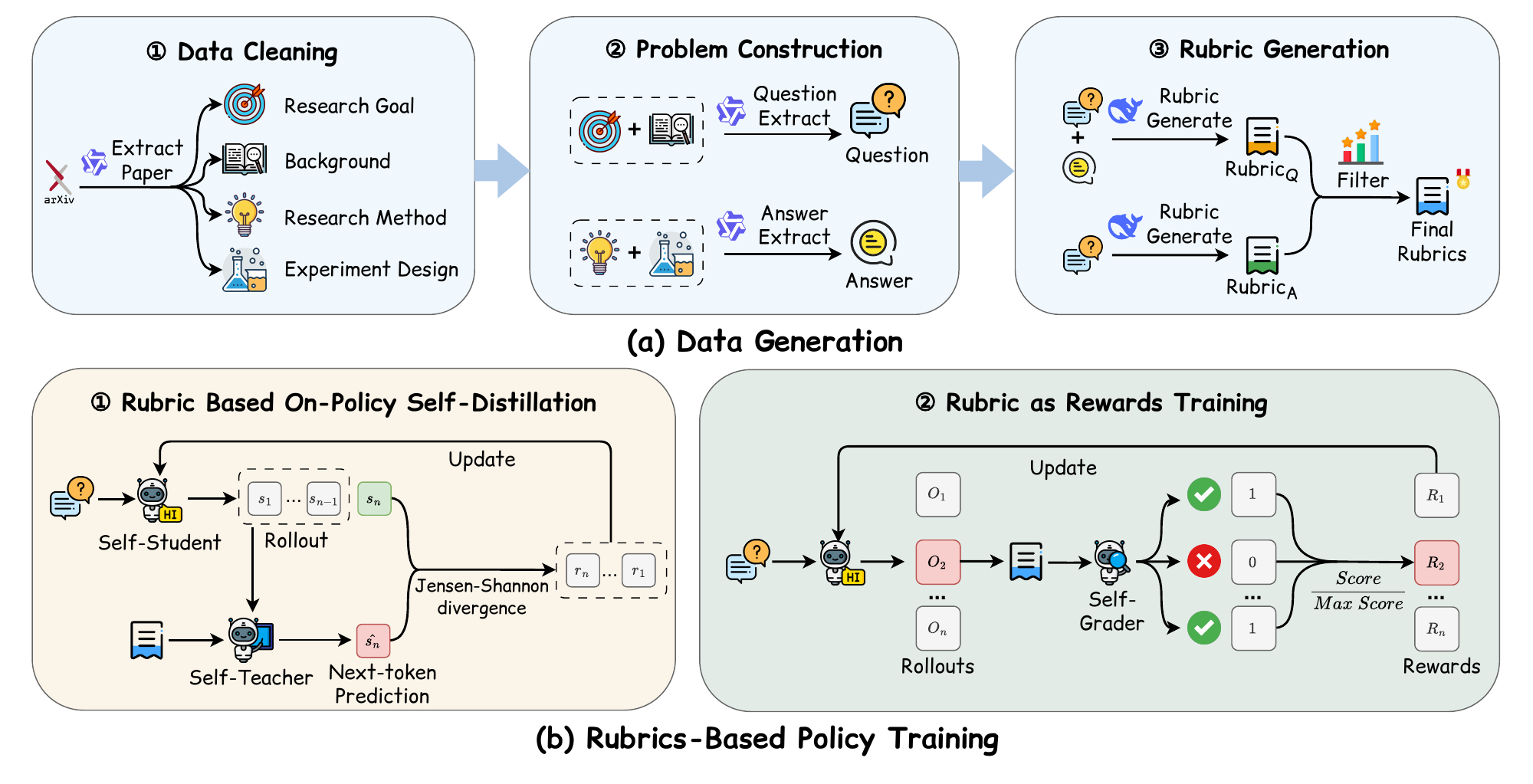}
    \caption{\textbf{Overview of the \methodname framework.} (a) Data Generation: arXiv papers are parsed into four stages to synthesize questions and answers; rubrics are generated, merged, ranked, and filtered. (b) Two-Stage Policy Training: rubric-based OPSD followed by rubric-as-rewards GRPO.}
    \label{fig:framework}
\end{figure*}

We formalize research-plan generation as learning a policy $\pi_\theta(a \mid q)$ that produces a coherent solution $a$ given a research question $q$. We turn each paper into a complete training environment whose task is the question and whose critic is the rubric. To address the two weaknesses of existing rubric-based pipelines identified in the introduction, our framework draws the question and the criteria from disjoint paper sections to eliminate criterion leakage, and consumes the rubric twice in a two-stage evolution, first as privileged context for a self-distillation teacher and then as the reward for GRPO, to recover the dense supervision that scalar rewards discard. The framework thus decomposes quality into independently verifiable criteria via a structured Rubric (Figure~\ref{fig:framework}), comprising a data generation stage and a rubric-based training stage.

\subsection{Dataset Construction}

\subsubsection{Data Preprocessing}

We build our dataset from publicly available arXiv papers, extracting plain-text LaTeX source for cleaner content than rendered PDFs. Each paper is decomposed into four stages — Research Goal, Background, Research Method, and Experimental Design — where the latter excludes concrete numerical results to prevent spurious memorization. The decomposition follows a map-reduce procedure. In the \emph{map} step, we split the paper into its natural sections and call Qwen3-235B-A22B to extract relevant information for each stage per section. In the \emph{reduce} step, we group extractions by stage and merge duplicates, yielding a coherent four-stage summary per paper (\Secref{sec:appendix-extraction}).

Given this representation, we frame the training task as open-ended question answering: the research question is synthesized from the Research Goal and Background, and the reference answer from the Research Method and Experimental Design. Because question and answer are drawn from disjoint paper sections, this construction prevents criterion leakage at its source.

\subsubsection{Rubric Generation}

The rubrics in our framework comprise two complementary parts. The specialized rubrics $\mathcal{R}_{\text{spec}}$ are instance-specific criteria tailored to each research question, measuring how well the response addresses the particular scientific problem. The general rubrics $\mathcal{R}_{\text{gen}}$ are fixed criteria applied uniformly across all instances, enforcing the completeness, specificity, and soundness of the proposed plan.

\paragraph{Specialized rubrics.} For each instance, we prompt DeepSeek-V4-Flash to generate $m$ criteria along two dimensions (methodological innovation and experimental design) from two complementary sources: question-conditioned rubrics $\mathcal{R}_Q$, derived from the research question alone, and answer-grounded rubrics $\mathcal{R}_A$, derived from both the question and the reference answer (\Secref{sec:appendix-qa}). We merge $\mathcal{R}_Q$ and $\mathcal{R}_A$, deduplicate semantically overlapping criteria, rank the survivors by importance, and retain the top $n$ criteria per instance ($m{=}n{=}10$). Because the reference answer is drawn from disjoint sections, the answer-grounded candidates encode content the question alone cannot reveal; ranking survivors by importance against this answer retains their core methodological and experimental-design decisions, grounding the final rubric in the method and experiments, consistent with the low measured leakage (Table~\ref{tab:leakage}).

\paragraph{General rubrics.} For instance-agnostic criteria, we adopt the general rubric of \citet{goel2025training}, which draws on prior work on research idea assessment~\citep{si2025can} and rubric-based benchmarks such as HealthBench~\citep{arora2025healthbench}, covering completeness, specificity, soundness, efficiency, and ethical safety (\Secref{sec:appendix-reward}). Reusing a validated rubric avoids evaluation bias from ad-hoc design.

\begin{figure}[t]
    \centering
    \includegraphics[width=\textwidth]{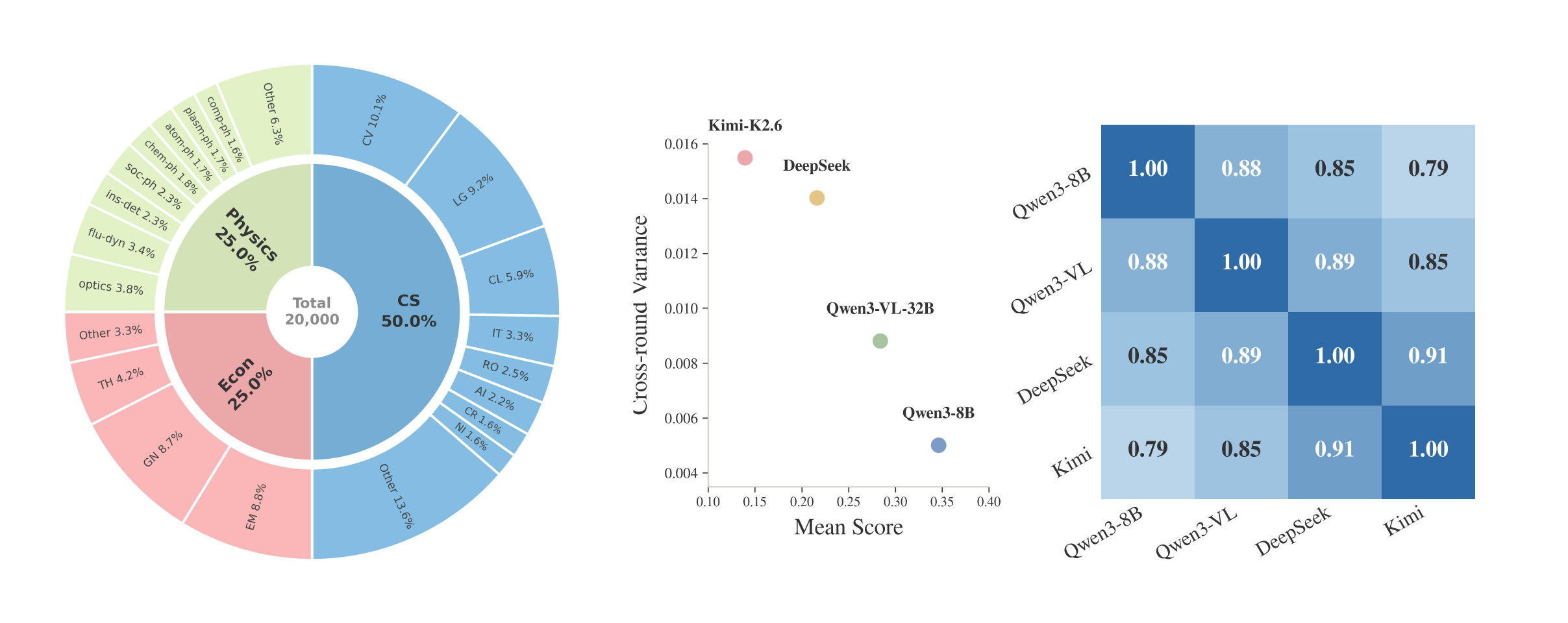}
    \caption{\textbf{Preliminary analysis.} \textbf{Left:} Category breakdown of \methodname-20k across three domains. \textbf{Middle:} Mean score vs.~across-round range for four scoring models over five independent runs. \textbf{Right:} Pairwise inter-model agreement on binary rubric verdicts.}
    \label{fig:data_analysis}
\end{figure}

\paragraph{Data analysis.} The final dataset comprises 20{,}000 instances across three domains, with CS the largest ($\approx$50\%), followed by Physics ($\approx$25\%) and Econ ($\approx$25\%). Among the specialized rubrics, 63.8\% of criteria target Method and 36.2\% target Experiment (\Secref{sec:appendix-rubric-type}). The overall composition is illustrated in Figure~\ref{fig:data_analysis}.

To verify scorer reliability, we evaluate self-consistency and inter-model agreement. Four models score across five runs at temperature 0. As shown in Figure~\ref{fig:data_analysis}, all models are highly self-consistent, with stronger judges scoring more strictly. The small Qwen3-8B reaches nearly 80\% agreement with Kimi K2.6.

With the question and the criteria in place, each paper constitutes a complete training environment in which the question defines the task and the rubric supplies the critic that scores any candidate plan.

\subsection{Rubric-Centered Training}

To validate the quality of the framework, we train policies on it for research-plan generation. Both stages are supervised by the same rubrics and together form a two-stage, rubric-centered evolution paradigm.

\paragraph{Rubric-Conditioned OPSD.} We apply On-Policy Self-Distillation (OPSD)~\citep{zhao2026self} to build a broad prior for research-proposal generation. The student generates on-policy rollouts, while the rubric-conditioned teacher evaluates the same rollout prefixes with the rubrics as privileged information withheld at inference; the student matches the teacher by minimizing their KL divergence:
\begin{equation}
    \mathcal{L}_{\text{OPSD}}(\theta) = \mathbb{E}_{(x, \mathcal{R}) \sim \mathcal{D}} \left[ \mathbb{E}_{\hat{y} \sim \pi_{\theta}(\cdot \mid x)} \frac{1}{|\hat{y}|} \sum_{n=1}^{|\hat{y}|} \mathrm{JSD}_{\beta}\!\left( \mathrm{sg}\!\left( \pi_{\theta}(\cdot \mid x, \mathcal{R}, \hat{y}_{<n}) \right) \,\|\, \pi_{\theta}(\cdot \mid x, \hat{y}_{<n}) \right) \right],
    \label{eq:opsd}
\end{equation}
where $\mathrm{sg}$ denotes the stop-gradient operator. Moreover, the rubric enumerates the full set of quality criteria rather than one realized solution. Because it specifies evaluation principles rather than a single target completion, the teacher preserves a broader set of valid continuations than answer-conditioned distillation (\Secref{sec:appendix-opsd-privileged}), while supplying abstract principles aligned with the second-stage reward.

\paragraph{GRPO with Rubric-as-Rewards.} In the second stage, we align the policy with Group Relative Policy Optimization (GRPO)~\citep{shao2024deepseekmath} against rubric-based rewards. The reward is produced by a frozen copy of the base model, which performs self-grading: conditioned on the research question, the reference answer, and the rubric, it grades each candidate response criterion by criterion and emits binary verdicts parsed into per-criterion scores, eliminating the need for a separately trained external reward model (\Secref{sec:appendix-reward}). The per-set scores $r_{i,\text{spec}}$ and $r_{i,\text{gen}}$ average the binary criterion verdicts over the specialized and general criterion sets $\mathcal{R}_{\text{spec}}$ and $\mathcal{R}_{\text{gen}}$, respectively, and the reward $r_i$ of a response $o_i$ combines the two as
\begin{equation}
    r_i = \alpha \, r_{i,\text{spec}} + (1-\alpha)\, r_{i,\text{gen}},
    \label{eq:reward}
\end{equation}
and we set $\alpha = 0.7$ in all experiments to place greater emphasis on task-specific scientific fit while maintaining nonzero pressure on general proposal quality, a choice validated by the sensitivity analysis in \Secref{sec:appendix-sensitivity}. We then optimize the policy with the GRPO objective
\begin{multline}
    \mathcal{J}_{\text{GRPO}}(\theta) = \mathbb{E}_{\{o_i\}_{i=1}^{G} \sim \pi_{\theta_{\mathrm{old}}}(\cdot \mid x)} \left[ \frac{1}{G} \sum_{i=1}^{G} \Big( \min\!\left( \rho_i A_i,\; \operatorname{clip}(\rho_i,\; 1-\varepsilon,\; 1+\varepsilon)\, A_i \right) \right. \\
    \left. {} - \beta\, D_{\mathrm{KL}}\!\left( \pi_{\theta}(\cdot \mid x) \,\|\, \pi_{\mathrm{ref}}(\cdot \mid x) \right) \Big) \right],
    \label{eq:grpo}
\end{multline}
where $\rho_i = \pi_{\theta}(o_i \mid x)/\pi_{\theta_{\mathrm{old}}}(o_i \mid x)$ is the sequence-level probability ratio of the $i$-th response and $A_i$ is its group-normalized advantage, obtained by standardizing the rubric-based reward $r_i$ of~\eqref{eq:reward} against the mean and standard deviation of the group rewards; the KL penalty to the reference policy $\pi_{\mathrm{ref}}$ prevents excessive drift.

\begin{table}[t]
    \centering
    \caption{Main results across model scales and training strategies. Subscripts denote changes relative to the corresponding base model. The best result in each section per column is in \textbf{bold}.}
    \label{tab:main_experiment}
    \resizebox{\textwidth}{!}{%
    \begin{tabular}{lcccccc}
        \toprule
        & \textbf{RubricHub} & \textbf{ResearchPlan} & \multirow{2}{*}{\textbf{ResearchQA}} & \multicolumn{2}{c}{\textbf{\methodname{}}} & \multirow{2}{*}{\textbf{Avg}} \\
        \cmidrule(lr){5-6}
        & Science & \textbf{-Gen} ML & & Innov & Design & \\
        \midrule
        \rowcolor{gray!10}
    \multicolumn{7}{c}{\textit{Proprietary Models}} \\
        Qwen3.5-Plus       & 63.74 & 47.32 & 75.12 & 27.28 & 27.07 & 48.11 \\
        GLM-5.1             & \textbf{63.93} & 51.89 & 78.69 & 30.44 & 28.73 & 50.74 \\
        Kimi K2.6          & 63.31 & 54.49 & 73.19 & 31.12 & 36.07 & 51.64 \\
        Claude-Sonnet-5    & 60.10 & 60.03 & 66.04 & 36.17 & \textbf{39.63} & 52.39 \\
        DeepSeek-V4-Flash  & 62.04 & 52.97 & \textbf{79.59} & 29.82 & 37.74 & 52.43 \\
        GPT-5.1            & 63.75 & \textbf{61.66} & 76.52 & \textbf{36.59} & 34.11 & \textbf{54.53} \\
        \midrule
        \rowcolor{gray!10}
    \multicolumn{7}{c}{\textit{Rubric-based Models}} \\
        Intern-S1-Mini    & 42.69 & 18.43 & 68.60 & 7.48  & 4.48 & 28.34 \\
        Rebicon-Preview    & 56.83 & 25.40 & \textbf{76.59} & 20.88 & 20.42 & 40.02 \\
        S1-VL-RL          & \textbf{57.77} & \textbf{29.23} & 72.09 & \textbf{22.31} & \textbf{23.68} & \textbf{41.02} \\
        \midrule
        \rowcolor{gray!10}
    \multicolumn{7}{c}{\textit{\methodname{}-20k Trained Models}} \\
        Qwen3-1.7B     & 31.84 & 4.89  & 55.14 & 10.76 & 8.58 & 22.24 \\
        \quad w/ SFT                    & 31.95\dtp{+0.11} & 5.85\dtp{+0.96}  & 57.36\dtp{+2.22} & 11.88\dtp{+1.12} & 8.50\dtn{-0.08} & 23.11\dtp{+0.87} \\
        \quad w/ OPSD                   & 34.31\dtp{+2.47} & 6.75\dtp{+1.86}  & 60.61\dtp{+5.47} & 14.16\dtp{+3.40} & 10.45\dtp{+1.87} & 25.26\dtp{+3.01} \\
        \quad w/ GRPO                   & 31.62\dtn{-0.22} & 5.34\dtp{+0.45}  & 55.67\dtp{+0.53} & 11.47\dtp{+0.71} & 8.60\dtp{+0.02} & 22.54\dtp{+0.30} \\
        \quad w/ OPSD + GRPO              & \textbf{34.90}\dtp{+3.06} & \textbf{8.37}\dtp{+3.48}  & \textbf{65.93}\dtp{+10.79} & \textbf{17.38}\dtp{+6.62} & \textbf{12.45}\dtp{+3.87} & \textbf{27.81}\dtp{+5.56} \\
        \midrule
        Qwen3-4B       & 41.97 & 13.48 & 63.15 & 16.12 & 13.89 & 29.72 \\
        \quad w/ SFT                    & 41.57\dtn{-0.40} & 12.14\dtn{-1.34} & 64.87\dtp{+1.72} & 17.90\dtp{+1.78} & 15.04\dtp{+1.15} & 30.30\dtp{+0.58} \\
        \quad w/ OPSD                   & 44.03\dtp{+2.06} & 15.71\dtp{+2.23} & 66.93\dtp{+3.78} & 18.98\dtp{+2.86} & 15.06\dtp{+1.17} & 32.14\dtp{+2.42} \\
        \quad w/ GRPO                   & 43.64\dtp{+1.67} & 13.60\dtp{+0.12} & 66.97\dtp{+3.82} & 18.57\dtp{+2.45} & 15.95\dtp{+2.06} & 31.75\dtp{+2.02} \\
        \quad w/ OPSD + GRPO              & \textbf{45.15}\dtp{+3.18} & \textbf{16.93}\dtp{+3.45} & \textbf{70.74}\dtp{+7.59} & \textbf{22.81}\dtp{+6.69} & \textbf{18.19}\dtp{+4.30} & \textbf{34.76}\dtp{+5.04} \\
        \midrule
        Qwen3-8B       & 46.07 & 20.28 & 66.65 & 18.69 & 17.03 & 33.74 \\
        \quad w/ SFT                    & 45.29\dtn{-0.78} & 17.91\dtn{-2.37} & 67.53\dtp{+0.88} & 17.52\dtn{-1.17} & 15.68\dtn{-1.35} & 32.79\dtn{-0.96} \\
        \quad w/ OPSD                   & 47.41\dtp{+1.34} & 23.17\dtp{+2.89} & 69.17\dtp{+2.52} & 21.42\dtp{+2.73} & 18.04\dtp{+1.01} & 35.84\dtp{+2.10} \\
        \quad w/ GRPO                   & 48.09\dtp{+2.02} & 20.76\dtp{+0.48} & 71.02\dtp{+4.37} & 20.26\dtp{+1.57} & 19.17\dtp{+2.14} & 35.86\dtp{+2.12} \\
        \quad w/ OPSD + GRPO              & \textbf{49.41}\dtp{+3.34} & \textbf{23.53}\dtp{+3.25} & \textbf{73.48}\dtp{+6.83} & \textbf{24.47}\dtp{+5.78} & \textbf{21.88}\dtp{+4.85} & \textbf{38.55}\dtp{+4.81} \\
    
        \bottomrule
    \end{tabular}%
    }
\end{table}

\section{Experiments}

\subsection{Experimental Setup}
\label{sec:experimental-setup}

\paragraph{Datasets and Benchmarks.}
\textit{(1) Training data.} We train on the computer-science subset of \methodname{}-20k, which comprises $\sim$10{,}000 instances derived from arXiv papers published between January 2015 and December 2025.
\textit{(2) In-domain benchmarks.} We construct held-out benchmarks from papers released after January 2026 across three domains (cs, econ, physics), sampling 400 papers per domain with two test instances each: \methodname{}-Innov derives the question from Research Goal and Background, uses Research Method as the reference answer, and scores along the innovation dimension; \methodname{}-Design additionally incorporates Research Method into the question, uses Experimental Design as the reference answer, and scores along the design dimension (\Secref{sec:appendix-testset}).
\textit{(3) Out-of-domain benchmarks.} Generalization is assessed on ResearchQA~\citep{yifei2025researchqa} (750 instances, 10 per field), ResearchPlanGen-ML~\citep{goel2025training} (full test set), and RubricHub Science~\citep{li2026rubrichub} (800 instances). All responses are scored by DeepSeek-V4-Flash via an LLM-as-a-judge protocol that rates rubric coverage per criterion.

\paragraph{Baselines.} We compare against two categories of baselines. The first consists of training-strategy baselines applied at every Qwen3 scale~\citep{yang2025qwen3}: the untrained base model, supervised fine-tuning (SFT) on the reference answers, and the single-stage OPSD and GRPO variants. The second consists of reference models evaluated without any fine-tuning on our data, spanning the general-purpose models Qwen3.5-Plus~\citep{qwen2026qwen35}, DeepSeek-V4-Flash~\citep{xu2026deepseek}, GLM-5.1~\citep{zeng2026glm}, Kimi K2.6~\citep{kimi2025k26}, GPT-5.1~\citep{singh2025openai} and Claude-Sonnet-5~\citep{anthropic2026sonnet5}, together with three models specialized for scientific tasks, Intern-S1-Mini (8B)~\citep{bai2025intern}, Rebicon-Preview (30B-A3B)~\citep{huang2025reinforcement}, and S1-VL-RL (32B)~\citep{li2026s1}.

\paragraph{Implementation Details.} We train three model scales from the Qwen3 family: Qwen3-1.7B and Qwen3-4B on four NVIDIA A6000 GPUs with 48 GB each, and Qwen3-8B on four NVIDIA Pro A6000 GPUs with 96 GB each. All models are used with thinking mode enabled by default during both training and evaluation. Both training stages use bf16 precision. In the OPSD stage, we fine-tune with LoRA ($r=64$, $\alpha=128$) at a learning rate of $5\times10^{-6}$ and an effective batch size of 8, training for 200 steps on the 1.7B and 4B models and 400 steps on the 8B model. In the GRPO stage, we train with the verl framework, sampling 8 responses per prompt via vLLM with a KL penalty of 0.01 and a learning rate of $1\times10^{-5}$, for 200 steps on the 1.7B and 4B models and 100 steps on the 8B model. For rubric scoring, the 4B and 8B models serve as their own verifiers, whereas the 1.7B model, whose scoring ability is insufficient, is scored by the 4B model.

\subsection{Main Results}

We present the results in three parts: the gain of the two-stage schedule within a fixed model family, the comparison with existing specialized and general-purpose models, and a win-rate study that isolates the contribution of the training data.

\paragraph{Within-Model Training Strategies.} Table~\ref{tab:main_experiment} presents the main results across model scales and training strategies. Relative to the untrained Qwen3 models, supervised fine-tuning (SFT) on the reference answers yields only marginal or even negative gains (dropping by $-0.96$ on average at the 8B scale). SFT merely imitates the single reference answer, collapsing the output distribution into a narrow mode and thereby failing to instill rubric alignment. In contrast, OPSD aligns the model's output distribution with a rubric-conditioned teacher while preserving output diversity: at minimal training cost, it matches single-stage GRPO performance (e.g., 35.84 vs.\ 35.86 average at the 8B scale, Table~\ref{tab:main_experiment}), thereby providing a stable initial strategy. Its entropy dynamics explain why this warm-up helps (Figure~\ref{fig:entropy}): entropy rises steadily during OPSD as new knowledge is absorbed, yielding a broad, well-structured prior on which the second-stage GRPO converges sharply to complete the final policy optimization. Accordingly, the OPSD+GRPO scheme achieves the highest score on every test set at every scale, improving the average score by $+5.56$, $+5.04$, and $+4.81$ on Qwen3-1.7B, 4B, and 8B, respectively, confirming that the OPSD warm-up promotes rapid convergence of the subsequent GRPO stage and that the two stages are complementary.

\paragraph{Comparison with Existing Models.}  Our rubric-centered training pipeline produces highly competitive models against both specialized scientific-task models and mainstream general-purpose LLMs. On the in-domain benchmarks, the fine-tuned Qwen3-8B outperforms the scientific-task baselines Intern-S1-Mini and Rebicon-Preview on both \methodname{}-Innov (24.47 vs.\ 7.48 and 20.88) and \methodname{}-Design (21.88 vs.\ 4.48 and 20.42), and also surpasses S1-VL-RL on ResearchQA (73.48 vs.\ 72.09). Among general-purpose LLMs, the same model is highly competitive on ResearchQA, even surpassing Kimi K2.6 (73.48 vs.\ 73.19). These results demonstrate that our rubric-centered data construction and the two-stage evolution training paradigm are highly competitive across diverse scientific tasks.

\paragraph{Win Rate.} Table~\ref{tab:three_model_win_rate} reports the three-way win rates of the baseline trained on \methodname{}-20k, the baseline trained on RubricHub Science, and the untrained base model on ResearchPlanGen-ML, a benchmark independent of all training data, following the judge protocol in \Secref{sec:appendix-winrate-three}. The model trained on \methodname{}-20k achieves the best performance across all dimensions, winning 58.1\% of Overall Score comparisons, compared with 28.2\% for its RubricHub-trained counterpart and 13.7\% for the base model. Since all three models share the same training regimen, these margins highlight the value of our rubric-centered data construction.

\subsection{Analysis}

\begin{table*}[t]
\centering
\caption{\textbf{Win Rate Results.} For each criterion, we report the percentage of comparisons in which each model is selected as the best among the baseline trained on \colorbox{oursyellow}{\methodname{}-20k}, the baseline trained on \colorbox{rubrichubteal}{RubricHub Science}, and the untrained \colorbox{baseblue}{\textcolor{white}{Qwen3-1.7B}} by an LLM judge. }
\label{tab:three_model_win_rate}

\newbox\cmpcellbox
\newbox\cmpfirstbox
\newcommand{\winrateimg}[2]{%
  \setbox\cmpcellbox=\hbox{\begin{tabular}{>{\raggedright\arraybackslash}p{8.0cm}}#1\end{tabular}}%
  \setbox\cmpfirstbox=\hbox{\parbox[t]{8.0cm}{\raggedright #1}}%
  \raisebox{\dimexpr -0.5\height + 0.5\ht\cmpfirstbox + 0.5\baselineskip - 0.5\dp\strutbox - 0.5\dp\cmpcellbox\relax}%
    {\includegraphics[width=0.35\textwidth]{#2}}%
}
\newcommand{\overallDesc}{Overall scientific quality and practical usefulness as a research blueprint.}
\newcommand{\goalDesc}{Understanding and addressing the core research goal, including identifying key challenges and formulating relevant hypotheses.}
\newcommand{\novelDesc}{Originality and research insight, including novel hypotheses, perspectives, or methodologies connected to the research goal.}
\newcommand{\soundDesc}{Scientific rigor and coherence, from hypothesis to methodology to evidence, with consideration of baselines, controls, and validation.}
\newcommand{\execDesc}{Clarity, feasibility, and implementation structure, balancing ambition with practicality.}
\newcommand{\impactDesc}{Likelihood of generating meaningful and interpretable research outcomes that advance understanding or provide methodological contributions.}

\resizebox{\textwidth}{!}{%
\begin{tabular}{>{\raggedright\arraybackslash}p{2.5cm} >{\raggedright\arraybackslash}p{9.5cm} c}
\toprule
\textbf{Criteria} & \textbf{Description} & \textbf{Win Rate} \\
\midrule
\textbf{Overall Score} & \overallDesc & \winrateimg{\overallDesc}{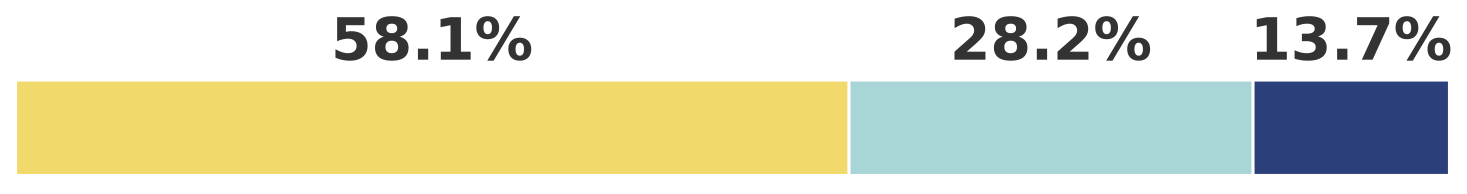} \\
\midrule
\addlinespace
Goal Alignment & \goalDesc & \winrateimg{\goalDesc}{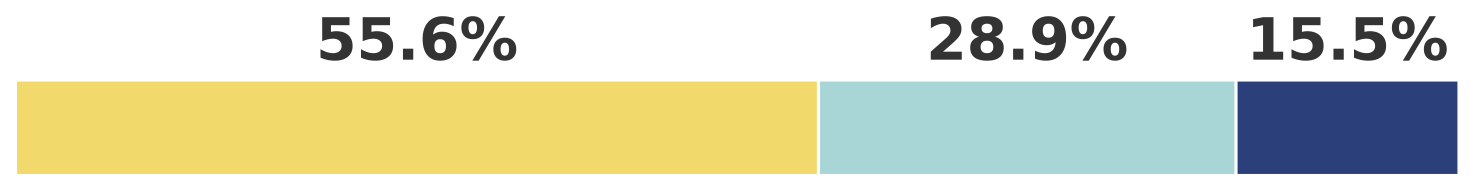} \\
\addlinespace
Novel Insight & \novelDesc & \winrateimg{\novelDesc}{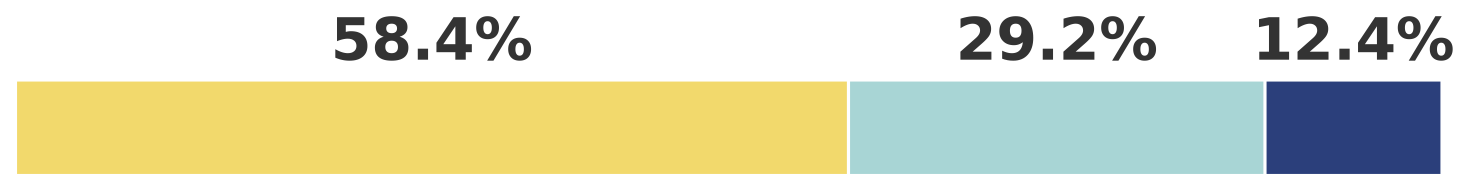} \\
\addlinespace
Scientific Soundness & \soundDesc & \winrateimg{\soundDesc}{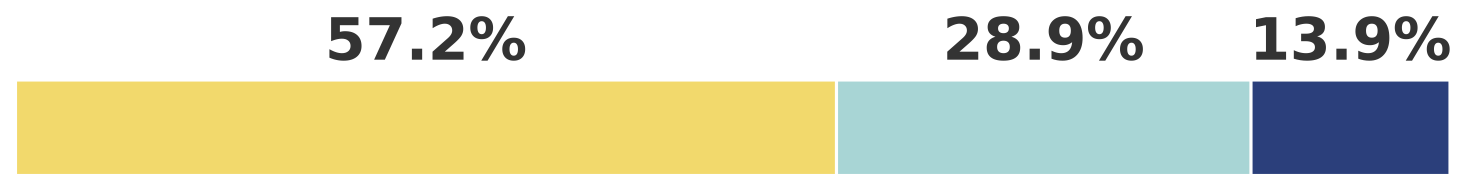} \\
\addlinespace
Execution Quality & \execDesc & \winrateimg{\execDesc}{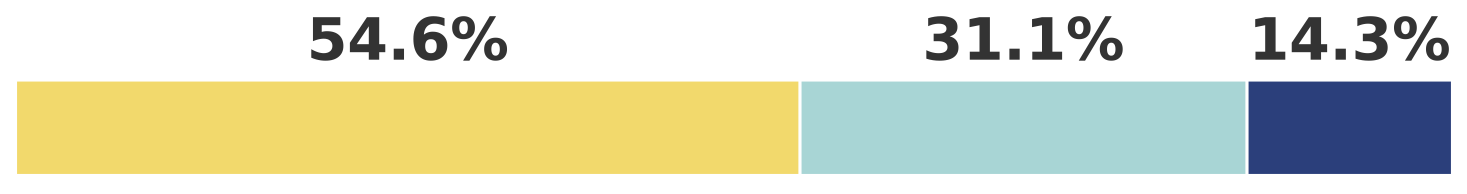} \\
\addlinespace
Expected Impact & \impactDesc & \winrateimg{\impactDesc}{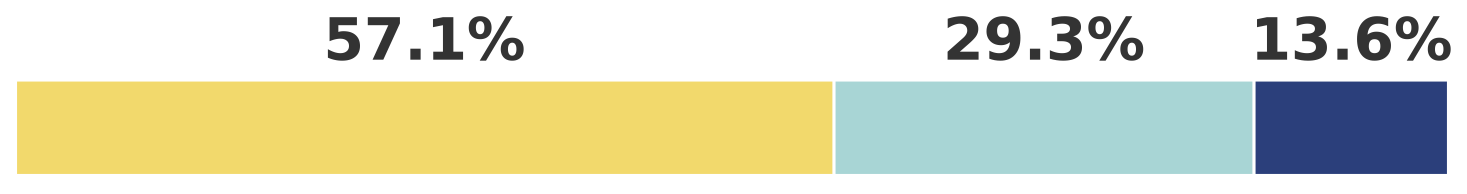} \\
\bottomrule
\end{tabular}%
}
\end{table*}

\subsubsection{Scaling Laws}

\begin{figure}[htbp]
    \centering
    \includegraphics[width=\textwidth]{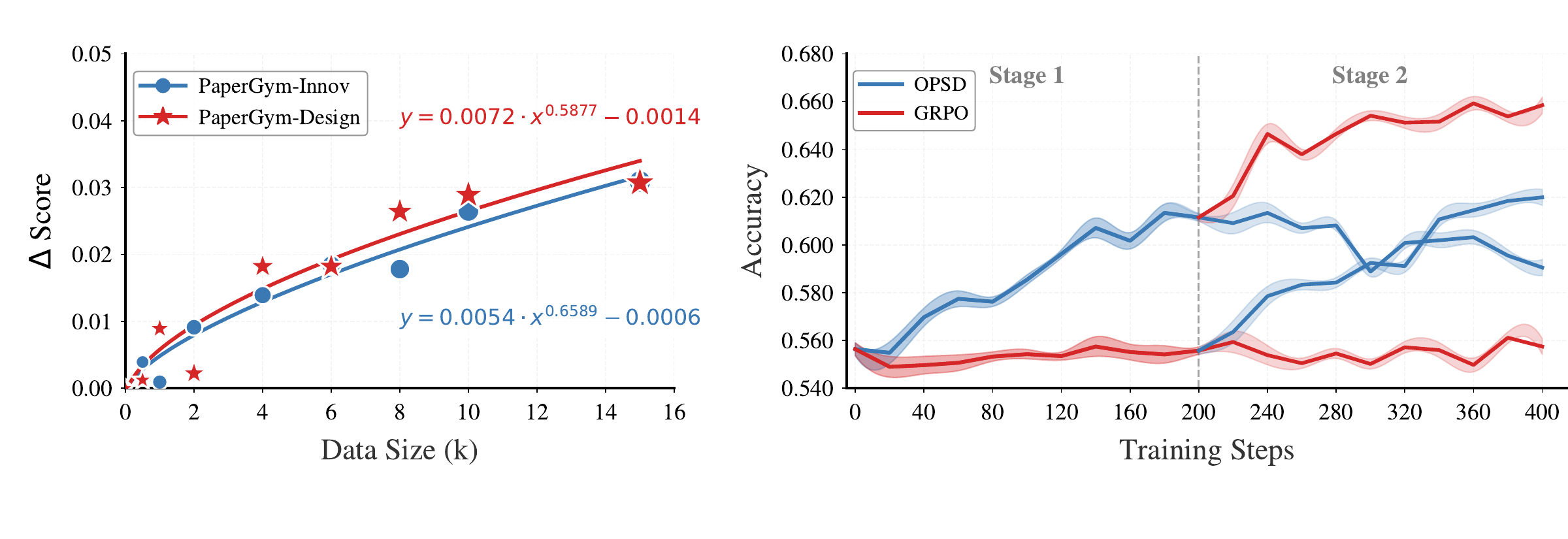}
    \caption{\textbf{Left:} Data scaling law with GRPO training. \textbf{Right:} Training dynamics across two-stage orderings of OPSD and GRPO on Qwen3-1.7B.}
    \label{fig:scale_and_train}
\end{figure}

To further verify the quality of our framework, we examine how model performance scales with the amount of training data. We train Qwen3-4B with GRPO using a batch size of 50, a group size of 8, and a learning rate of $1\times10^{-5}$, and progressively increase the dataset from 0.5k to 15k instances by sampling evenly across the three domains of \methodname{}-20k (5k instances per domain at the largest scale). As shown in Figure~\ref{fig:scale_and_train} (Left), performance improves with data scale: the trained model advances from 16.12 to 19.21 on \methodname{}-Innov and from 13.89 to 16.96 on \methodname{}-Design, both exhibiting a clear growth trend. This confirms a positive correlation between data quantity and model capability and further attests to the quality of our framework.

\subsubsection{Criterion Leakage}

To quantify the criterion leakage of our decoupling pipeline, we present DeepSeek-V4-Flash with each research question and its associated rubric and ask whether each criterion can be directly inferred from the question alone (\Secref{sec:appendix-leakage}). As shown in Table~\ref{tab:leakage}, the four-stage decomposition suppresses spurious correlations between the input and the target, keeping the leakage rate at only 3.73\%, 4.71\%, and 4.97\% on \methodname{}-20k, \methodname{}-Innov, \methodname{}-Design, respectively, compared with 11.90\%--34.10\% for existing benchmarks. This roughly 3--9$\times$ reduction establishes a clean foundation for subsequent training.

\subsection{Ablations}

\subsubsection{Ablation on Rubric}

To isolate the contribution of each design choice in our data-construction pipeline, we evaluate six configurations that vary rubric source, dimension, and model capability (Table~\ref{tab:data_ablation}), each trained by fine-tuning Qwen3-1.7B with OPSD for 200 steps on the resulting dataset. The full pipeline achieves the best overall performance, confirming the value of combining both rubric sources and dimensions.

\begin{table}[t]
    \centering
    \label{tab:leakage_and_ablation}
    \begin{minipage}[t]{0.35\textwidth}
        \centering
        \caption{Criterion leakage comparison.}
        \label{tab:leakage}
        \begin{tabular}{lc}
            \toprule
             & \textbf{Leakage Rate} \\
            \midrule
            \methodname{}-20k    & \textbf{3.73\%} \\
            \midrule
            \methodname{}-Innov    &  4.71\% \\
            \methodname{}-Design   &  4.97\% \\
            \midrule
            HealthBench           & 11.90\% \\
            RubricHub Science     & 17.39\% \\
            ResearchPlanGen-ML    & 31.29\% \\
            ResearchPlanGen-ArXiv & 34.10\% \\
            ResearchQA            & 19.22\% \\
            \bottomrule
        \end{tabular}
    \end{minipage}
    \hfill
    \begin{minipage}[t]{0.50\textwidth}
        \centering
        \caption{Ablation on data construction pipeline.}
        \label{tab:data_ablation}
        \begin{tabular}{lcc}
            \toprule
            & \textbf{Innov} & \textbf{Design} \\
            \midrule
            Full Pipeline (Ours) & \textbf{14.16} & \textbf{10.45} \\
            \midrule
            \rowcolor{gray!10}
            \multicolumn{3}{l}{\textit{Rubric Generation}} \\
            \quad $R_Q$-only           & 13.19 & 10.24 \\
            \quad $R_A$-only           & 12.29 & 9.34 \\
            \quad Innovation dim.\ only & 14.13 & 9.93 \\
            \midrule
            \rowcolor{gray!10}
            \multicolumn{3}{l}{\textit{Model Scale}} \\
            \quad Extract $\to$ Qwen3-8B & 13.19 & 9.78 \\
            \quad Rubric $\to$ Qwen3-8B  & 12.16 & 9.27 \\
            \bottomrule
        \end{tabular}
    \end{minipage}
\end{table}

\paragraph{Rubric Generation.} The full pipeline outperforms all rubric-related variants. Both single-source variants fall below the full pipeline: $R_Q$-only by -0.97 / -0.21 and $R_A$-only by -1.87 / -1.11, confirming that neither rubric source alone provides sufficient supervision and that integrating the two is necessary. Restricting the rubric to the innovation dimension alone (-0.03 / -0.52) likewise underperforms, validating our complementary dual-dimensional design.

\paragraph{Model Scale.} Using Qwen3-8B for extraction causes moderate drops (-0.97 / -0.67), while replacing the rubric generator causes sharper drops (-2.00 / -1.18). This contrast indicates that rubric quality is the critical bottleneck: high-quality rubrics demand nuanced scientific judgment where the gap between strong and weak models is most consequential, whereas the extraction pipeline is inherently robust.

\subsubsection{Ablation on Training Stage}
\label{sec:ablation-training-stage}

To study the interplay between OPSD and GRPO, Figure~\ref{fig:scale_and_train} (Right) plots the ResearchQA accuracy over 400 training steps for Qwen3-1.7B under four strategies: OPSD alone, GRPO alone, OPSD+GRPO, and GRPO+OPSD. These runs are separate from Table~\ref{tab:main_experiment}; each strategy is trained for 400 steps to show the full trajectory, which is why the OPSD-only curve extends past its peak.

\paragraph{Stage Ordering.} OPSD alone peaks around step 200 and then declines because its privileged information drives rapid early progress but overfits, capping performance without GRPO refinement. GRPO alone starts behind OPSD because exploration from the base policy is unstable, an effect most pronounced at smaller scales. In contrast, OPSD+GRPO consistently outperforms all alternatives, including GRPO+OPSD, by converging faster to a higher final accuracy. The key is the warm-up: OPSD structures the initial distribution for GRPO to refine, whereas applying GRPO first on random init yields unstable exploration and worse outcomes. This advantage extends to all scales (\Secref{sec:appendix-stage-order}).

\begin{figure}[htbp]
    \centering
    \includegraphics[width=\textwidth]{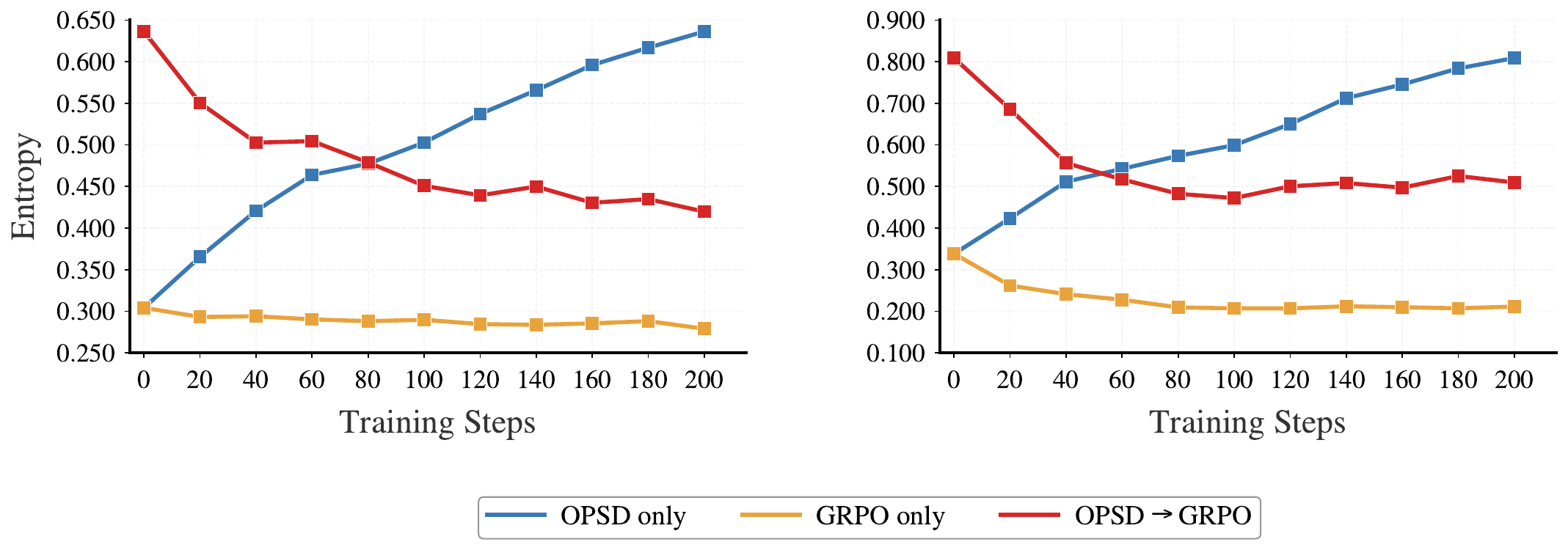}
    \caption{Entropy dynamics under different training strategies.}
    \label{fig:entropy}
\end{figure}

\paragraph{Entropy Dynamics.} The entropy trajectories during training explain the advantage of OPSD+GRPO, as OPSD raises output entropy by injecting new knowledge, whereas GRPO lowers it by converging toward high-reward regions. Crucially, GRPO's entropy drop is sharper after the OPSD warm-up, confirming that OPSD provides a broader, better-structured prior for GRPO to refine (Figure~\ref{fig:entropy}; details in \Secref{sec:appendix-entropy}).

\section{Conclusion}

We presented \methodname{}, a unified framework that turns each research paper into a complete training environment for research-plan generation. Its four-stage extraction pipeline synthesizes the research question from the research goal and background, and the reference solution from the research method and experimental design, decoupling inputs from answers and eliminating criterion leakage at the source. The resulting 20{,}000-instance corpus \methodname{}-20k exhibits a criterion-leakage rate of 3.7\%, the lowest among open-source alternatives. Instance-specific rubrics examine both methodological innovation and experimental design, and training uses the same rubric twice: first as privileged context for a self-distillation teacher, then as the reward for GRPO. Qwen3 models trained with this rubric-centered evolution improve consistently on in-domain and external benchmarks, and the trained 8B model surpasses the far larger Kimi K2.6 on ResearchQA.

\clearpage

\bibliography{main,custom}
\bibliographystyle{main}

\appendix
\tableofcontents
\clearpage

\appendix

\textbf{Note on the main results.} In Table~\ref{tab:main_experiment}, Claude-Sonnet-5 scores notably lower on ResearchQA (66.04) than the other general-purpose models. This underperformance stems from the model's refusal behavior: on a subset of ResearchQA questions, Claude-Sonnet-5 judges that the information it has retrieved is insufficient to fully answer the question and therefore declines to answer. Because unanswered questions receive a score of zero, this behavior lowers the model's final accuracy; the reported number thus reflects this refusal behavior rather than a deficiency in research-plan generation ability.

\section{Dataset Analysis}

\subsection{Rubric Type Distribution}
\label{sec:appendix-rubric-type}
To measure the distribution of Method-oriented and Experiment-oriented criteria in our specialized rubrics, we randomly sample 1{,}000 questions from the training set and ask DeepSeek-V4-Flash to classify every criterion of the sampled instances as either \emph{Method} or \emph{Experiment}. \emph{Method} captures criteria concerning the novelty of the proposed method, algorithm, model, or framework itself, whereas \emph{Experiment} captures criteria concerning experimental setup, evaluation metrics, baseline comparisons, ablation studies, dataset selection, and other experiment-related designs. Each criterion is classified independently given the research question and the full rubric of that instance. The classification prompt is as follows:

\begin{promptbox}
\begin{Verbatim}[breaklines=true, breakanywhere=false]
You are an expert reviewer of academic papers. For each rubric item below, classify it as either "Method" or "Experiment" based on these definitions:

1. **Method**: Involves proposing a new method, algorithm, model, framework, technique, theoretical contribution, or novel aspects of the solution design itself.
2. **Experiment**: Involves experimental setup, evaluation metrics, baseline comparisons, ablation studies, dataset selection, human-machine comparison experiments, cross-institution/cross-user comparisons, robustness verification, and other experiment-related designs.

Output only one label per line, one for each rubric, in the same order as listed below, and nothing else.

Problem context:
{problem_context}

Rubrics:
{rubrics_text}

Output:
\end{Verbatim}
\end{promptbox}

Across the 1{,}000 sampled instances, 9{,}999 criteria are classified in total, of which 6{,}375 (63.76\%) are \emph{Method} and 3{,}624 (36.24\%) are \emph{Experiment}. This indicates that our specialized rubrics emphasize methodological innovation while still devoting substantial weight to experimental design.

\subsection{Criterion Leakage Detection}
\label{sec:appendix-leakage}
We quantify criterion leakage using an LLM-based judge. For each instance, we feed the research question together with all of its rubric criteria to DeepSeek-V4-Flash in a single call and ask it to output, for every criterion, a binary verdict on whether that criterion can be directly inferred from the question alone. The leakage rate of a dataset is the fraction of criteria marked as directly inferable. The detection prompt is as follows:

\begin{promptbox}
\begin{Verbatim}[breaklines=true, breakanywhere=false]
You are an expert in evaluating dataset quality and detecting criterion leakage.

For each rubric item below, determine whether it can be directly inferred from the question alone.
Output only "Yes" or "No" for each rubric item, one per line, and nothing else.

Question: {question}

Rubrics:
{rubrics_text}

Output:
\end{Verbatim}
\end{promptbox}

We apply this detection to both our in-domain benchmarks and a diverse set of out-of-domain benchmarks, including HealthBench, RubricHub Science, ResearchPlanGen-ML, ResearchPlanGen-ArXiv, and ResearchQA. The resulting leakage rates, reported in Table~\ref{tab:leakage}, confirm that our decoupling pipeline attains a substantially lower leakage rate than existing benchmarks.

\section{Training Details and Additional Ablations}

\subsection{Privileged-Information Selection for OPSD}
\label{sec:appendix-opsd-privileged}
The OPSD teacher can be conditioned on different forms of privileged information. To justify our choice of rubrics, we ablate the privileged information while keeping all other settings fixed: starting from the base model Qwen3-1.7B, we train with OPSD for 200 steps on \methodname{}-20k and compare three variants in which the teacher conditions on (i) the rubric (our default), (ii) the reference answer, and (iii) the rubric together with the reference answer. Table~\ref{tab:opsd_privileged_ablation} reports the resulting rubric scores on the two in-domain benchmarks \methodname{}-Innov and \methodname{}-Design.

\begin{table}
    \centering
    \caption{Ablation on the privileged information provided to the teacher in the OPSD stage. Best results are in \textbf{bold}.}
    \label{tab:opsd_privileged_ablation}
    \begin{tabular}{lcc}
        \toprule
        Privileged Information & Innov & Design \\
        \midrule
        Rubric (Ours)             & \textbf{14.16} & \textbf{10.45} \\
        Reference Answer          & 13.08 & 9.57 \\
        Rubric + Reference Answer & 13.81 & 9.17 \\
        \bottomrule
    \end{tabular}
\end{table}

The rubric-conditioned teacher achieves the best score on both in-domain benchmarks, outperforming the answer-conditioned teacher by +1.08 on Innov and +0.88 on Design. Because the reference answer specifies a single realized solution, answer-conditioned distillation narrows the set of valid continuations that the teacher can preserve, so the distilled prior transfers less to downstream research-plan generation. Conditioning on the rubric in addition to the reference answer does not recover this gap and even further degrades the Design score (9.17 vs.~10.45 with the rubric alone), indicating that the single realized solution conflicts with the broader principle-level supervision. These results confirm that, on the research-plan generation task, the rubric provides better privileged information than the reference answer and substantiate our choice of rubric-conditioned OPSD in Eq.~\eqref{eq:opsd}.

\subsection{Parameter Sensitivity}
\label{sec:appendix-sensitivity}
To justify the mixing weight $\alpha$ in Eq.~\eqref{eq:reward}, we sweep the ratio of the specialized to the general rubric term while keeping all other settings fixed. Starting from the Qwen3-1.7B model trained with 200 OPSD steps, we run the GRPO stage for 100 steps with the specialized-to-general ratio set to $8{:}2$, $7{:}3$, and $6{:}4$, and evaluate the resulting policies on the two in-domain benchmarks \methodname{}-Innov and \methodname{}-Design. Table~\ref{tab:sensitivity} reports the rubric scores. The $7{:}3$ configuration attains the highest score on both benchmarks, $17.17$ on \methodname{}-Innov and $12.35$ on \methodname{}-Design, confirming $\alpha=0.7$ as the optimal setting.

\begin{table}
    \centering
    \caption{Sensitivity of the GRPO stage to the reward mixing ratio (specialized : general). Best results are in \textbf{bold}.}
    \label{tab:sensitivity}
    \begin{tabular}{lcc}
        \toprule
        Specialized : General & Innov & Design \\
        \midrule
        $8{:}2$  & 14.63 & 10.86 \\
        $7{:}3$  & \textbf{17.17} & \textbf{12.35} \\
        $6{:}4$  & 16.40 & 11.54 \\
        \bottomrule
    \end{tabular}
\end{table}

\subsection{Stage-Ordering Swap between OPSD and GRPO}
\label{sec:appendix-stage-order}
While the main-text ablation (\Secref{sec:ablation-training-stage}) compares the two orderings at the 1.7B scale, here we quantify the effect of the swap across all three scales and five benchmarks. We train the reverse ordering GRPO$\rightarrow$OPSD with all settings identical to the default schedule except the stage order: the same data and hyperparameters, with 200 GRPO and 200 OPSD steps for the 1.7B and 4B models, and 100 GRPO and 400 OPSD steps for the 8B model.

\begin{figure}[htbp]
    \centering
    \includegraphics[width=0.7\textwidth]{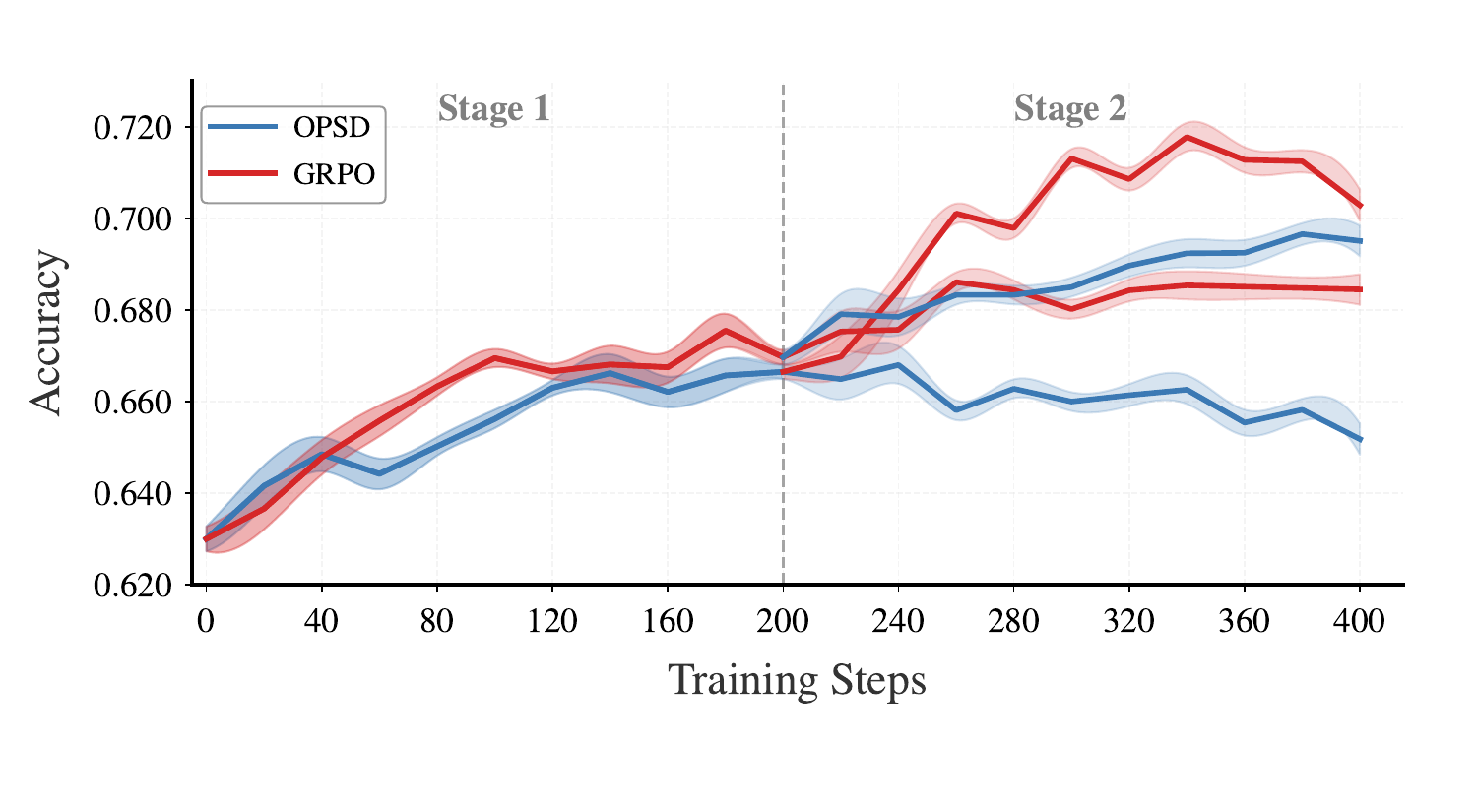}
    \caption{Training dynamics across the two stage orderings of OPSD and GRPO on Qwen3-4B, the 4B counterpart of the right panel of Figure~\ref{fig:scale_and_train}. OPSD$\rightarrow$GRPO leads throughout training and converges to a higher final accuracy than GRPO$\rightarrow$OPSD.}
    \label{fig:scale_and_train_4b}
\end{figure}

Table~\ref{tab:stage_order} reports the benchmark scores of both orderings. The forward ordering wins on \emph{every} benchmark at \emph{every} scale, improving the five-benchmark average by $+2.04$, $+0.99$, and $+1.31$ points on Qwen3-1.7B, 4B, and 8B; the penalty is largest at the 1.7B scale and consistently concentrates on \methodname{}-Innov ($+2.97$, $+2.27$, $+1.92$), the benchmark that most depends on retaining a broad prior over valid plans. Figure~\ref{fig:scale_and_train_4b} shows the 4B training dynamics, which replicate the 1.7B pattern.

We attribute the advantage of the forward ordering to two complementary mechanisms.

\paragraph{Cold-start instability of GRPO.} Run from a cold start, GRPO's on-policy rollouts are uniformly weak, so its group-normalized advantages (Eq.~\eqref{eq:grpo}) are dominated by sampling noise rather than by meaningful differences in plan quality, and the KL penalty simultaneously resists the large distribution shift required to escape the low-reward region; a large fraction of the GRPO budget is spent on unstable exploration. After an OPSD warm-up, the policy already produces rubric-aligned plans, group rewards become informative, and GRPO refines efficiently. Reversing the schedule forces GRPO to grapple with the cold-start instability that OPSD bypasses, while placing OPSD after GRPO yields little extra benefit—reward optimization already injects the rubric knowledge, and ending on distillation rather than on direct verification is inherently weaker.

\paragraph{A widen-then-narrow entropy curriculum.} Because OPSD broadens the output distribution while GRPO narrows it (see \Secref{sec:appendix-entropy}), the forward ordering realizes a beneficial \emph{broaden-then-optimize} curriculum, whereas the reverse ordering collapses entropy prematurely: GRPO narrows the distribution around whatever the cold-start policy discovers early, and the subsequent self-distillation cannot restore the lost diversity, since teacher and student are the same already-collapsed policy and the pruned modes no longer carry probability mass to be broadened.

\begin{table}[t]
    \centering
    \caption{Benchmark performance under the two stage orderings of the two-stage schedule. \emph{OPSD$\rightarrow$GRPO} is the default ordering; \emph{GRPO$\rightarrow$OPSD} swaps the two stages with all other settings identical. Subscripts on the \emph{GRPO$\rightarrow$OPSD} row denote the drop relative to \emph{OPSD$\rightarrow$GRPO}. The default ordering wins on every benchmark at every scale.}
    \label{tab:stage_order}
    \resizebox{\textwidth}{!}{%
    \begin{tabular}{lccccccc}
        \toprule
        & & \textbf{RubricHub} & \textbf{ResearchPlan} & \multirow{2}{*}{\textbf{ResearchQA}} & \multicolumn{2}{c}{\textbf{\methodname{}}} & \multirow{2}{*}{\textbf{Avg}} \\
        \cmidrule(lr){6-7}
        & & Science & \textbf{-Gen} ML & & Innov & Design & \\
        \midrule
        \multirow{2}{*}{Qwen3-1.7B}
            & OPSD$\rightarrow$GRPO  & \textbf{34.90} & \textbf{8.37} & \textbf{65.93} & \textbf{17.38} & \textbf{12.45} & \textbf{27.81} \\
            & GRPO$\rightarrow$OPSD & 34.09\dtn{-0.81} & 7.16\dtn{-1.21} & 61.86\dtn{-4.07} & 14.41\dtn{-2.97} & 11.35\dtn{-1.10} & 25.77\dtn{-2.04} \\
        \midrule
        \multirow{2}{*}{Qwen3-4B}
            & OPSD$\rightarrow$GRPO  & \textbf{45.15} & \textbf{16.93} & \textbf{70.74} & \textbf{22.81} & \textbf{18.19} & \textbf{34.76} \\
            & GRPO$\rightarrow$OPSD & 44.34\dtn{-0.81} & 16.30\dtn{-0.63} & 70.03\dtn{-0.71} & 20.54\dtn{-2.27} & 17.64\dtn{-0.55} & 33.77\dtn{-0.99} \\
        \midrule
        \multirow{2}{*}{Qwen3-8B}
            & OPSD$\rightarrow$GRPO  & \textbf{49.41} & \textbf{23.53} & \textbf{73.48} & \textbf{24.47} & \textbf{21.88} & \textbf{38.55} \\
            & GRPO$\rightarrow$OPSD & 48.36\dtn{-1.05} & 22.27\dtn{-1.26} & 72.53\dtn{-0.95} & 22.55\dtn{-1.92} & 20.53\dtn{-1.35} & 37.25\dtn{-1.31} \\
        \bottomrule
    \end{tabular}%
    }
\end{table}

\subsection{Reward Scoring Protocol}
\label{sec:appendix-reward}
During GRPO training, each candidate response is scored by a frozen copy of the base policy model, which acts as its own verifier; for the 1.7B model, whose self-scoring is unreliable, the 4B model serves as the scorer. The scorer is conditioned on the research question, the instance-specific rubric, and the reference answer, and grades each response in two independent passes: once against the instance-specific (specialized) rubric and once against seven general criteria that check whether the plan handles all stated criteria, provides a detailed and specific solution, leaves no overlooked flaws, is well justified, is cost- and effort-efficient, raises no ethical issues, and remains consistent with the overall plan. For every criterion the scorer emits a strict binary verdict: 1 only if the answer completely satisfies the criterion with quality comparable to or better than the reference answer, and 0 otherwise, with zero leniency for partially correct or ``on the right track'' responses. The binary verdicts are averaged with equal weights to obtain a per-pass score in $[0,1]$, and the final reward combines the specialized and general scores as $0.7\,R_{\text{spec}}+0.3\,R_{\text{gen}}$, matching Eq.~\eqref{eq:reward} with $\alpha=0.7$. All scorings are collected at temperature $0$ for determinism.

The seven general criteria, applied verbatim and uniformly to every instance, are listed in full below:

\begin{enumerate}[leftmargin=*]
    \item \textbf{Handles All Criteria}: Does the plan satisfy all criteria mentioned in the rubric item? An exception is if the criteria says `such as', `for example', or `including', the response does not have to include the same examples listed to meet the criteria, but whatever is provided must be valid and reasonable.
    \item \textbf{Detailed, Specific Solution}: Does the part of the plan relevant to satisfying this rubric item include fully specified details on HOW to implement it? There should be no self-proclaimed claims of handling something without doing so. There should be no vague terms, ambiguity, or lack of clarity. It should be described in simple to understand language.
    \item \textbf{No Overlooked Flaws or Weaknesses}: Are there any important overlooked flaws or weaknesses in the part of the plan addressing this rubric item that invalidate its satisfaction of the rubric item?
    \item \textbf{Well Justified Rationale}: Is the part of the plan relevant to this grading item well-motivated and justified? For example, are there convincing arguments provided for how the plan handles this grading item is better than simpler solutions or alternate hypotheses?
    \item \textbf{Cost and Effort Efficient}: Does the plan handle this rubric item cost efficiently, without being unnecessarily complex? Check if a solution requiring less human effort or resources would be equally effective for this rubric item.
    \item \textbf{No Ethical Issues}: Does this part of the plan have any potential for negative consequences, or is it ethically problematic?
    \item \textbf{Consistent with Overall Plan}: Is this part of the plan consistent with the overall approach?
\end{enumerate}

The scoring prompt is as follows:

\begin{promptbox}
\begin{Verbatim}[breaklines=true, breakanywhere=false]
You are an exceptionally rigorous and professional academic peer reviewer and domain expert. Your task is to conduct an objective, impartial, and high-standard evaluation of complex scientific research tasks, which may include methodology design, experimental setup, or data analysis pipelines.

### Evaluation Task
You must evaluate the provided answer against the given scoring rubrics. For each rubric, you must assign a strict binary score (0 or 1) indicating whether the answer completely satisfies the criterion.

### Input Data
<Question>
{question}
</Question>

<Rubrics>
{rubrics_text}
</Rubrics>

<Answer>
{answer}
</Answer>

<Reference_Answer>
{reference_answer}
</Reference_Answer>

### Strict Scoring Guidelines (MUST Follow):
1. **Score 1 (Completely Satisfies)**: The answer fully, explicitly, and comprehensively meets all requirements of the specific rubric, demonstrating quality comparable to or better than the reference answer.
2. **Score 0 (Does Not Satisfy)**: Assign 0 if ANY of the following apply:
   - The answer only partially meets the criterion or exhibits obvious methodological, experimental, or analytical flaws compared to the reference answer.
   - The content is vague, generic, or lacks specific technical details that are present in the reference answer.
   - Important elements required by the rubric are omitted.
   - The answer contains factual errors, logical fallacies, or scientific inaccuracies not present in the reference answer.
   - The response is too brief to demonstrate a complete research methodology compared to the reference answer.
   - The answer fails to address aspects of the rubric that the reference answer addresses properly.
3. **Comparison with Reference**: Use the reference answer as a benchmark for quality and completeness. The answer should meet the rubric at least as well as the reference answer does.
4. **ZERO Leniency**: Do NOT award a 1 simply because the answer is "on the right track" or "mostly correct". Strict precision is required for a score of 1.

### Output Requirements:
- You MUST output ONLY a valid JSON array. Do not output any thinking process or conversational filler.
- **CRITICAL**: You must provide the `reason` FIRST, and then the `score`.
- **IMPORTANT**: The number of output items in the JSON array MUST match the number of input rubrics EXACTLY. You have {num_rubrics} rubrics, so you must output exactly {num_rubrics} items. Do NOT use a fixed number from the example below - the example shows only 2-3 items for illustration, but you must output one item for EACH rubric listed above.

### Output Format:
Please wrap your output in ```json ``` blocks exactly like this:
```json
[
    {
        "rubric_id": 1,
        "reason": "Detailed explanation of why the answer fails to meet the standard, pointing out specific flaws or omissions compared to the reference answer.",
        "score": 0
    },
    {
        "rubric_id": 2,
        "reason": "Detailed explanation of exactly how the answer completely meets all requirements of this standard, comparing favorably with the reference answer.",
        "score": 1
    },
    {
        "rubric_id": 3,
        "reason": "...",
        "score": ...
    }
]
```
\end{Verbatim}
\end{promptbox}

\subsection{Entropy Dynamics}
\label{sec:appendix-entropy}
We attribute the consistent advantage of the two-stage ordering to the entropy dynamics observed during training (Figure~\ref{fig:entropy}). During the OPSD stage, entropy rises steadily, reflecting the injection of new knowledge through self-distillation over the full output distribution. In contrast, GRPO drives entropy reduction as the policy converges toward high-reward regions by selecting relatively optimal trajectories within each group. Crucially, the entropy drop during GRPO is substantially more pronounced when the model has been pre-warmed with OPSD. This indicates that OPSD establishes a broader and better-structured prior, which enables GRPO to perform more focused and efficient policy convergence.

\section{Win Rate Comparison}

\subsection{Pairwise Win-Rate Evaluation Protocol}
\label{sec:appendix-winrate-pairwise}
To further verify the quality of our framework and the capability gains from training, we conduct pairwise win-rate comparisons among four Qwen3-1.7B settings: (i) the untrained base model; (ii) 200 OPSD steps on the computer-science subset of \methodname{}-20k; (iii) 200 OPSD steps on RubricHub Science under identical OPSD hyperparameters; and (iv) the full two-stage schedule, i.e., 200 OPSD steps followed by 200 GRPO steps on the computer-science subset of \methodname{}-20k.

\begin{figure}[htbp]
    \centering
    \includegraphics[width=\textwidth]{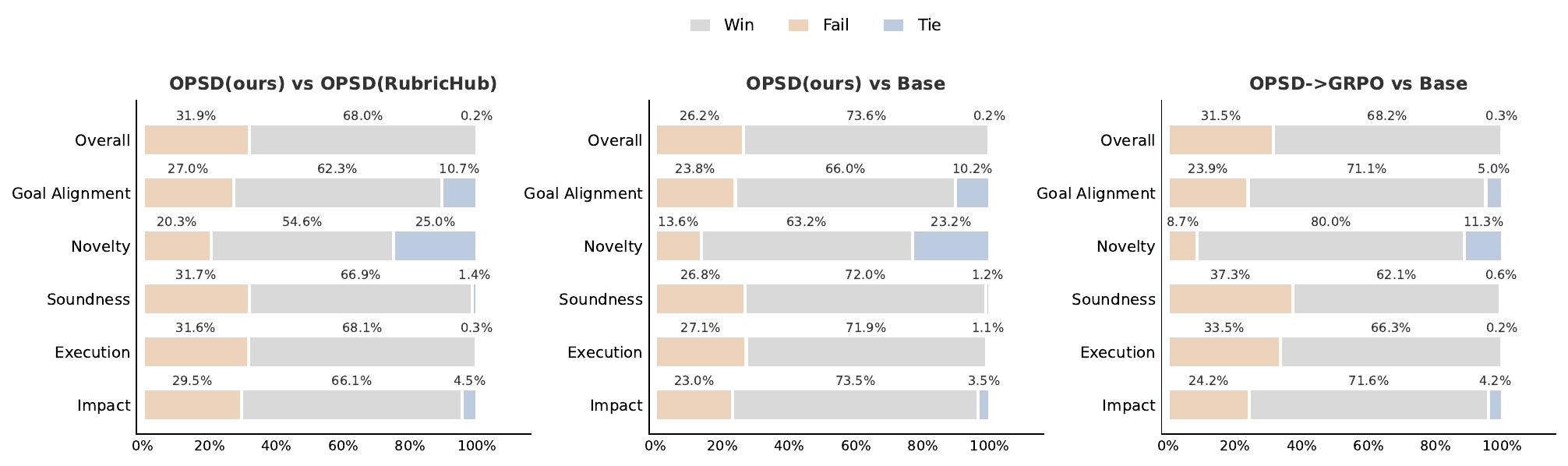}
    \caption{Pairwise win rates of fine-tuned Qwen3-1.7B variants on ResearchPlanGen-ML. Models trained on our data outperform the RubricHub-trained counterpart, and both trained models defeat the untrained base model, evidencing benchmark quality gains and genuine training improvements.}
    \label{fig:win_rate_detail}
\end{figure}

Figure~\ref{fig:win_rate_detail} reports the pairwise win rates on ResearchPlanGen-ML, a benchmark independent of all training data, scored by three expert judges (DeepSeek-V4-Flash, Kimi K2.6, and GLM-5.1). Our OPSD model defeats the RubricHub-trained OPSD model by 68.0\% overall, confirming the higher quality of our benchmark, while both our OPSD and two-stage models defeat the untrained base model (73.6\% and 68.2\% overall), confirming genuine capability gains from training.

However, GRPO in the second stage improves the model unevenly across criteria. Against the same base model, the two-stage model gains on goal alignment (66.0\% to 71.1\%) and novel insight (63.2\% to 80.0\%), but loses ground on scientific soundness (72.0\% to 62.1\%), execution quality (71.9\% to 66.3\%), and expected research impact (73.5\% to 71.6\%). We attribute this redistribution to two factors. First, methodological-innovation rubrics dominate our dataset, so innovation-oriented criteria receive both the densest reward signal and the largest headroom, since the model starts lowest there, explaining the gains on novel insight and goal alignment. Second, the model exhibits reward hacking, producing longer responses to match more rubric items, and the resulting increase in output complexity lowers the scores on the remaining criteria.

The pairwise evaluation prompt is structurally identical to the three-way prompt in \Secref{sec:appendix-winrate-three}, except that only two plans (Plan A and Plan B) are compared, and the judge selects the better plan or declares a tie for each criterion rather than ranking three plans.

\subsection{Three-Way Win-Rate Evaluation Protocol}
\label{sec:appendix-winrate-three}

To isolate the contribution of the training data, all three models are built from Qwen3-1.7B, and the two trained models share an identical 200-step OPSD recipe with the same hyperparameters as the main experiments (\Secref{sec:experimental-setup}): the untrained base model, the model trained on the computer-science subset of \methodname{}-20k, and the model trained on RubricHub Science. The only difference between the two trained models is the training corpus. These correspond to settings (i)--(iii) of the pairwise study in \Secref{sec:appendix-winrate-pairwise}.

We score all 685 instances in ResearchPlanGen-ML through three-way comparisons with an LLM-as-a-judge protocol under the supervision of three expert judges: DeepSeek-V4-Flash, Kimi K2.6, and GLM-5.1. For each research question, we collect one response from each of the three models under comparison and present the question together with the three responses to every judge. To eliminate positional bias, the three responses are randomly assigned to the labels Plan A, Plan B, and Plan C before being shown, and the judge is never told which label corresponds to which model. Each judge independently evaluates the three plans along five criteria (goal alignment, novel insight, scientific soundness, execution quality, and expected research impact) and ranks them from best to worst for each criterion. The judge also assigns each plan an overall score from 1 to 10 that reflects its value as a research blueprint.

Per-criterion rankings are aggregated across the three judges via majority voting: for each criterion, the model that receives the most votes (i.e., is ranked highest by the most judges) is declared the winner. When the three judges split their votes (one vote for each model), a fourth expert judge (Gemini 3.7 Flash) serves as a tie-breaker. The overall score of each model is averaged across the three judges. The win rate of a model is the fraction of questions on which it achieves the highest averaged overall score among the three. All judgments are collected at temperature 0 for determinism. The evaluation prompt is as follows:

\begin{promptbox}
\begin{Verbatim}[breaklines=true, breakanywhere=false]
You are tasked with comparing three research plans for the same research scenario. You need to evaluate them based on specific criteria and provide scores. The order they are provided here is randomized.

# Research Scenario
{scenario}

# Research Plan A
{plan_a}

# Research Plan B
{plan_b}

# Research Plan C
{plan_c}

# Evaluation Criteria
Evaluate each research plan based on the following criteria:

1. Goal Alignment
Which plan better understands and addresses the core research goal? A strong plan should identify the key scientific challenges, formulate relevant research questions or hypotheses, and avoid unnecessary experiments or directions that do not contribute to the stated objective.

2. Novel Insight
Which plan demonstrates stronger originality and research insight? A strong plan should propose novel hypotheses, perspectives, methodologies, or experimental designs rather than simply applying existing approaches. Novelty should be meaningful and connected to the research goal.

3. Scientific Soundness
Which plan provides a more rigorous and convincing scientific investigation? A strong plan should have a coherent reasoning chain from hypothesis to methodology to evidence. It should consider appropriate baselines, controls, confounding factors, alternative explanations, and experimental validation strategies.

4. Execution Quality
Which plan is clearer, more feasible, and better structured for implementation? A strong plan should specify what needs to be done, how it will be done, and why each step is necessary. It should balance ambition and practicality, avoiding unnecessary complexity while retaining essential experiments.

5. Expected Research Impact
Which plan is more likely to generate meaningful research outcomes if successfully executed? A strong plan should lead to interpretable findings, advance understanding of the research problem, or provide valuable methodological contributions.

# Overall Score

Finally, provide an overall score (1-10) for each research plan.

The score reflects how valuable the plan would be as a research blueprint if assigned to an average graduate student in the relevant field. Consider both scientific quality and practical usefulness.

Score guidelines:

10: Exceptional research plan.
The plan is highly insightful, rigorous, and well-designed. If executed as described, it would likely produce a strong research contribution with minimal additional guidance.

9: Excellent research plan.
The plan has a strong research idea, solid methodology, and clear execution path. Only minor improvements are needed.

8: Strong research plan.
The plan is coherent and useful, with good scientific reasoning. It may contain some limitations in novelty, rigor, or details, but can likely lead to a successful study with moderate refinement.

7: Good research plan.
The plan addresses the research goal appropriately and contains valuable ideas, but has noticeable weaknesses in methodology, clarity, feasibility, or depth.

6: Moderately useful research plan.
The plan provides a reasonable direction but lacks important details, has methodological gaps, or requires substantial additional insight to become a strong study.

5: Mixed-quality research plan.
The plan contains some relevant ideas but also significant weaknesses. It may be partially useful as a reference, but requires major improvement before execution.

4: Weak research plan.
The plan is related to the research goal but has substantial problems in reasoning, methodology, feasibility, or research focus.

3: Poor research plan.
The plan contains major flaws and would likely mislead a researcher if followed without significant correction.

2: Very poor research plan.
The plan is largely ineffective, with serious conceptual or methodological issues.

1: Invalid research plan.
The plan is irrelevant to the research goal, fundamentally incorrect, or unusable.

# Output Format
For each criterion, rank the three plans from best to worst (1 = best, 3 = worst). Ties are allowed.

Provide your analysis and judgment in the following XML format:
<evaluation>
    <reasoning>
    Think critically and skeptically about all three plans. Consider their weaknesses and strengths. Compare them relative to each other on each criterion.
    </reasoning>
    <criteria_judgments>
        <goal_alignment>Rank plans A, B, C from best to worst</goal_alignment>
        <novel_insight>Rank plans A, B, C from best to worst</novel_insight>
        <scientific_soundness>Rank plans A, B, C from best to worst</scientific_soundness>
        <execution_quality>Rank plans A, B, C from best to worst</execution_quality>
        <expected_impact>Rank plans A, B, C from best to worst</expected_impact>
    </criteria_judgments>
    <overall_score_a>Score from 1-10 for Plan A</overall_score_a>
    <overall_score_b>Score from 1-10 for Plan B</overall_score_b>
    <overall_score_c>Score from 1-10 for Plan C</overall_score_c>
</evaluation>
\end{Verbatim}
\end{promptbox}

\section{Data Construction Details}

\subsection{Four-Stage Knowledge Extraction}
\label{sec:appendix-extraction}
We document the prompts behind the map-reduce decomposition of each paper. In the map step, the LaTeX body is split along its section boundaries, and Qwen3-235B-A22B is prompted to extract, from each section and in a single call, the salient information pertaining to the four stages, namely the research field, the research background, the solution, and the experimental design. The prompt enforces strict faithfulness: extracted content must appear verbatim in the source section, categories with no sentence-level evidence are returned as empty strings, and LaTeX symbols in the output are double-escaped so that the JSON remains parseable. The map prompt, with \texttt{\{section\_content\}} denoting the current section, is as follows:

\begin{promptbox}
\begin{Verbatim}[breaklines=true, breakanywhere=false]
You are an expert academic researcher.
Analyze the following paper section and extract information according to four categories.

[Paper Section Content]:
{section_content}

[Four Categories to Extract]:
1. **research_field**: The specific academic discipline, sub-field, and focused research direction of the paper.
2. **research_background**: Current state of the field, historical/related work, and limitations/gaps of prior works.
3. **solution**: Core idea, motivation, methodology, technical means, model architecture, and key components proposed by the authors of THIS paper.
4. **experiment**: Experimental design methodology, datasets, baselines, evaluation metrics, implementation details, and analysis strategies.

[Critical Rules]:
1. **STRICTLY extract from original text**: You MUST only extract information that appears in the provided section content. DO NOT fabricate, infer, or add any content not explicitly present in the text. If you cannot find sentence-level evidence in the section, do not extract it.
2. For each category, if relevant information exists in the section, extract and concisely summarize the key points. If NO relevant information exists, set the field to an empty string "".
3. The extracted information should be in the original language of the paper.
4. You MUST output your response in STRICT, valid JSON format.
5. CRITICAL: If your output contains LaTeX mathematical symbols (e.g., \sum, \mathbb), you MUST double-escape the backslashes in the JSON string (e.g., use \\sum, \\mathbb).
6. Do not include any conversational text, reasoning, or thinking process.
7. ABSOLUTE PROHIBITION: DO NOT output <think> tags, step-by-step explanations, or any reasoning.

[Output Format]:
Please wrap your output in ```json ``` blocks exactly like this:
```json
{
    "research_field": "extracted research field info or empty string",
    "research_background": "extracted research background info or empty string",
    "solution": "extracted solution info or empty string",
    "experiment": "extracted experiment info or empty string"
}
```
\end{Verbatim}
\end{promptbox}

In the reduce step, the per-section extractions of each stage are grouped and fed back to Qwen3-235B-A22B to be merged into a coherent, non-redundant summary. Each stage uses a dedicated prompt, with \texttt{\{combined\_info\}} denoting the concatenated per-section extractions of that stage. The four reduce prompts are as follows:

\textbf{Research field.}
\begin{promptbox}
\begin{Verbatim}[breaklines=true, breakanywhere=false]
You are an expert academic taxonomy specialist.
Based on the extracted information from various sections of a paper, synthesize a highly concise description of its Research Field.
Extracted Information:
{combined_info}
Task Requirements:
1. Define the Broad Discipline (e.g., Computer Science, Biology).
2. Define the Specific Sub-field (e.g., Natural Language Processing, Reinforcement Learning).
3. Define the Focused Research Direction.
4. EXCLUDE any background context, proposed methods, or experimental details.

Please output a concise summary in 1-3 sentences directly. Do not use filler words like "The research field of this paper is...".
\end{Verbatim}
\end{promptbox}

\textbf{Research background.}
\begin{promptbox}
\begin{Verbatim}[breaklines=true, breakanywhere=false]
You are an expert academic paper reviewer.
Based on the following extracted information from various sections of a paper, comprehensively summarize the 'Research Background'.
Extracted Information:
{combined_info}
Task Requirements:
Please synthesize the information and structure your response strictly into the following three parts:
1. **Field Overview**: Briefly introduce the current state and background of the research field.
2. **Historical Work**: Summarize the existing methods, related works, and their characteristics.
3. **Core Problems (Gaps)**: Clearly identify the limitations of existing works and the specific problems this paper aims to solve.
Note: Focus ONLY on the background. Do not describe the specific novel solutions or experimental results proposed in this paper. Make the summary academic, coherent, and concise.
\end{Verbatim}
\end{promptbox}

\textbf{Solution.}
\begin{promptbox}
\begin{Verbatim}[breaklines=true, breakanywhere=false]
You are an expert academic paper reviewer.
Based on the following extracted methodological details from a paper, comprehensively describe the proposed 'Solution'.
Extracted Information:
{combined_info}
Task Requirements:
Please organize your detailed description strictly using the following 4 dimensions (Use Markdown formatting):

1. **Core Idea and Motivation**: Core problems addressed, basic idea of the method, and the primary motivation.
2. **Method Architecture and Technical Details**: Overall architecture, key technical means, algorithm flow, or mathematical principles.
3. **Key Technical Components**: Core components, their responsibilities, and how they collaborate.
4. **Core Innovation Points**: How this method innovates compared to existing/traditional approaches.
Notes:
- Faithfully summarize based ONLY on the provided content.
- Prioritize preserving specific technical details and key terminology.
- Exclude overly detailed numerical settings (e.g., learning rates, batch sizes, epochs) and focus on method-level design.
- If information is insufficient for a specific dimension, summarize what is available but ensure the core architecture is clear.
\end{Verbatim}
\end{promptbox}

\textbf{Experimental design.}
\begin{promptbox}
\begin{Verbatim}[breaklines=true, breakanywhere=false]
You are an expert academic paper reviewer.
Based on the following extracted experimental details from a paper, comprehensively describe the 'Experiment Design, Process, and Analysis Strategy'.
Extracted Information:
{combined_info}
Task Requirements:
Please organize your detailed description strictly using the following 5 dimensions (Use Markdown formatting):

1. **Experimental Design and Framework**: Overall experimental setup, methodology, and structural organization.
2. **Data and Baseline Configuration**: Dataset selection, preprocessing steps, baseline methods, and comparison protocols.
3. **Evaluation and Metrics**: Performance measures, evaluation criteria, and validation procedures.
4. **Implementation and Settings**: Parameter configurations, environment setup, and implementation details.
5. **Analysis and Validation Strategy**: Ablation studies, statistical validation, and component contribution analysis.
Notes:
- Faithfully summarize based ONLY on the provided content.
- Prioritize preserving specific experimental details and methodological information.
- Exclude raw numerical results, focusing on the experimental methodology and process.
- If information is insufficient for a specific dimension, summarize what is available but ensure the experimental approach is clear.
\end{Verbatim}
\end{promptbox}

\subsection{Question-Answer and Rubric Construction}
\label{sec:appendix-qa}
Given the four-stage summary, Qwen3-235B-A22B synthesizes the research question from the research field and background, and the reference answer from the solution and experimental-design stages, so that the question and answer are drawn from disjoint paper sections and no instance leaks its answer into its question. DeepSeek-V4-Flash then generates the specialized rubrics from two complementary sources and merges them. The question prompt is as follows:

\begin{promptbox}
\begin{Verbatim}[breaklines=true, breakanywhere=false]
# Role
You are an experienced researcher with expertise in formulating concise research problem statements.

# Task
Based on the provided [Research Field] and [Research Background], write a concise research problem statement as a single coherent paragraph. The statement must naturally convey:
- The specific unsolved problem, limitation, or gap that needs to be addressed
- The relevant discipline and sub-field
- The current state of existing approaches and key challenges

# Output Format
Wrap the final statement inside <output> </output> tags. Rules for the content inside the tags:
- Write as a single paragraph of fluent, flowing prose
- Do NOT use bullet points, numbered lists, or section headers of any kind
- Include NO introductory phrases (e.g., "Here is the statement:", "Based on the background:")
- Include NO conversational filler or meta-commentary (e.g., "I have written...", "This statement covers...")
- Output only the statement itself, nothing else

# Input
[Research Field]:
{research_field}

[Research Background]:
{research_background}
\end{Verbatim}
\end{promptbox}

The reference-answer prompt requires faithfulness to the source, excludes concrete numerical results, and describes the method and experimental design in formal academic prose:

\begin{promptbox}
\begin{Verbatim}[breaklines=true, breakanywhere=false]
# Role
You are an experienced researcher tasked with writing a reference solution for a research problem. Your writing must be rigorous, academic in style, logically structured, and concise.

# Task
Based on the provided [Research Problem], [Original Solution], and [Original Experiment], write a reference solution that describes the method and experimental design. The description must adhere to the following principles:

1. **Faithfulness**: Only describe content that is directly supported by the provided material. Do NOT add any fabricated details, speculative claims, or information not present in the inputs.
2. **No concrete experimental data**: Exclude all specific numerical results, statistics, or quantitative outcomes. Focus on the rationale and design logic of the method and experiments, not the raw numbers.
3. **Content focus**: Emphasize the principles and implementation of the proposed method, the motivation behind the experimental design (e.g., choice of baselines, datasets, evaluation metrics, variable controls), and the reasoning for the approach.
4. **Academic style**: Use formal, precise, and concise academic language. Write in a structured but paragraph-based flow - avoid bullet points, numbered lists, or section headers. Each sentence should convey a clear technical point.

# Output Format
Wrap the final reference solution inside <output> </output> tags. The content within the tags should be the solution only, without introductory phrases, conversational filler, or meta-commentary.

# Input
[Research Problem]:
{research_problem}

[Original Solution]:
{raw_solution}

[Original Experiment]:
{raw_experiment}
\end{Verbatim}
\end{promptbox}

The question-conditioned rubrics $\mathcal{R}_Q$ are derived from the research question alone, whereas the answer-grounded rubrics $\mathcal{R}_A$ additionally condition on the reference answer. In both cases the model is asked for exactly ten binary criteria across the two dimensions of methodological innovation and experimental design. The answer-grounded rubric prompt is as follows:

\begin{promptbox}
\begin{Verbatim}[breaklines=true, breakanywhere=false]
# Role
You are an expert in scientific methodology and solution evaluation. Your task is to analyze the given research problem and reference solution, and generate comprehensive evaluation rubrics to assess the quality of any proposed solution in a binary (True/False) manner.

# Research Problem
{research_problem}

# Reference Solution
{reference_solution}

# Rubric Design Guidelines
Each rubric must be:
- **True/False binary**: objectively assessable without subjective interpretation.
- **Atomic**: evaluate ONE and only ONE independent concept per rubric.
- **Objective**: avoid subjective adjectives (good, detailed, clear, reasonable); use deterministic phrasing (e.g., "Contains formula X", "Mentions concept Y").
- **Negation-style positively phrased**: e.g., "No logical conflicts are present" rather than "Logical conflicts occurred".
- **Not about final results**: do NOT include rubrics that require executing experiments to verify.
- **Not about trivial details**: avoid minor/overly specific details of the reference solution, as multiple valid approaches may exist.

Construct rubrics evenly across the following two dimensions:

1. **Method Principles and Implementation**: Assess whether the solution correctly describes the core principles of the method, the technical implementation logic, key procedural steps, necessary constraints, and practical feasibility.

2. **Experimental Design**: Assess whether the solution proposes a sound and well-justified experimental setup, including appropriate baselines, datasets, evaluation metrics, variable controls, handling of confounds, and reproducibility considerations.

Prioritize rubrics that target the **core innovation points and key steps** of the solution's method and experimental design - discard trivial or peripheral aspects. Fewer high-impact rubrics are better than many shallow ones.

# Output Requirements
Generate exactly 10 rubrics. Avoid overly specific numerical settings (e.g., learning rates, batch sizes, epochs). Provide strictly valid JSON.
Wrap the output JSON array in ```json ``` code blocks, formatted as:
```json
[
  {"id": 1, "description": "..."},
  {"id": 2, "description": "..."}
]
```
\end{Verbatim}
\end{promptbox}

The two rubric sets are merged by DeepSeek-V4-Flash, which deduplicates semantically overlapping criteria and ranks the survivors by importance against the reference answer, keeping the top ten as the final specialized rubric. The merge prompt, with \texttt{\{rubric\_set\_a\}} and \texttt{\{rubric\_set\_b\}} denoting the JSON-encoded description lists of the two candidate sets, is as follows:

\begin{promptbox}
\begin{Verbatim}[breaklines=true, breakanywhere=false]
# Role
You are an expert in scientific evaluation methodology. Your task is to synthesize a definitive set of evaluation rubrics by merging two candidate lists.

# Task
You are provided with a [Research Problem], a [Reference Solution], and two sets of binary rubrics ([Set A] and [Set B]). Perform the following:

1. **Strategic Merging & Deduplication**: Identify rubrics that assess the same underlying technical criterion. When duplicates or semantic overlaps occur:
    - Select the single best version from the existing candidates that is most specific and measurable.
    - **Crucial**: You must pick one of the original strings. Do NOT synthesize new wording or paraphrase.
    - Remove all other redundant versions.

2. **Importance Ranking**: Rank the unique rubrics from high to low importance based on the [Reference Solution]:
    - **Top Tier**: Rubrics assessing the core novelty, theoretical breakthroughs, or critical technical steps.
    - **Lower Tier**: Rubrics assessing routine setup, standard metrics, or auxiliary experimental details.

# Research Problem
{research_problem}

# Reference Solution
{reference_solution}

# Rubric Set A (Derived from Ground Truth)
{rubric_set_a}

# Rubric Set B (Derived from Problem Only)
{rubric_set_b}

# Output Requirements
- Output ONLY a valid JSON array of objects.
- Each object must follow this exact schema: {"description": "string"}.
- No other fields (id, weight, etc.) are allowed.
- No conversational filler, introductory text, or markdown formatting outside the JSON block.
- The order of elements in the array must reflect their importance (descending).

[
  {"description": "..."},
  ...
]
\end{Verbatim}
\end{promptbox}

\subsection{Test Set Construction}
\label{sec:appendix-testset}
The two in-domain benchmarks are produced by two variants of the same pipeline that isolate a single dimension. For \methodname{}-Innov, the research-question prompt is identical to the one above, and only the reference-answer and rubric prompts differ: the reference answer is written from the Research Method stage alone, and all rubrics are elicited along the methodological-innovation dimension only. The method-only reference-answer prompt is as follows:

\begin{promptbox}
\begin{Verbatim}[breaklines=true, breakanywhere=false]
# Role
You are an experienced researcher tasked with writing a reference solution for a research problem. Your writing must be rigorous, academic in style, logically structured, and concise.

# Task
Based on the provided [Research Problem] and [Original Solution], write a reference solution that describes the proposed method. The description must adhere to the following principles:

1. **Faithfulness**: Only describe content that is directly supported by the provided material. Do NOT add any fabricated details, speculative claims, or information not present in the inputs.
2. **No concrete experimental data**: Exclude all specific numerical results, statistics, or quantitative outcomes. Focus on the rationale and design logic of the method, not the raw numbers.
3. **Content focus**: Emphasize the principles and implementation of the proposed method, including the core technical innovation, key algorithmic steps, necessary constraints, and practical feasibility. Do NOT include experimental design details (e.g., baselines, datasets, evaluation metrics).
4. **Academic style**: Use formal, precise, and concise academic language. Write in a structured but paragraph-based flow - avoid bullet points, numbered lists, or section headers. Each sentence should convey a clear technical point.

# Output Format
Wrap the final reference solution inside <output> </output> tags. The content within the tags should be the solution only, without introductory phrases, conversational filler, or meta-commentary.

# Input
[Research Problem]:
{research_problem}

[Original Solution]:
{raw_solution}
\end{Verbatim}
\end{promptbox}

The method-only rubric prompt for \methodname{}-Innov restricts the dimension to method principles and implementation:

\begin{promptbox}
\begin{Verbatim}[breaklines=true, breakanywhere=false]
# Role
You are an expert in scientific methodology and solution evaluation. Your task is to analyze the given research problem and reference solution, and generate comprehensive evaluation rubrics to assess the quality of any proposed solution in a binary (True/False) manner.

# Research Problem
{research_problem}

# Reference Solution
{reference_solution}

# Rubric Design Guidelines
Each rubric must be:
- **True/False binary**: objectively assessable without subjective interpretation.
- **Atomic**: evaluate ONE and only ONE independent concept per rubric.
- **Objective**: avoid subjective adjectives (good, detailed, clear, reasonable); use deterministic phrasing (e.g., "Contains formula X", "Mentions concept Y").
- **Negation-style positively phrased**: e.g., "No logical conflicts are present" rather than "Logical conflicts occurred".
- **Not about final results**: do NOT include rubrics that require executing experiments to verify.
- **Not about trivial details**: avoid minor/overly specific details of the reference solution, as multiple valid approaches may exist.

Construct rubrics focusing exclusively on the following dimension:

**Method Principles and Implementation**: Assess whether the solution correctly describes the core principles of the method, the technical implementation logic, key procedural steps, necessary constraints, practical feasibility, and the novelty of the proposed approach. Do NOT include rubrics related to experimental design, baselines, datasets, or evaluation metrics.

Prioritize rubrics that target the **core innovation points and key steps** of the solution's method - discard trivial or peripheral aspects. Fewer high-impact rubrics are better than many shallow ones.

# Output Requirements
Generate exactly 10 rubrics. Avoid overly specific numerical settings (e.g., learning rates, batch sizes, epochs). Provide strictly valid JSON.
Wrap the output JSON array in ```json ``` code blocks, formatted as:
```json
[
  {"id": 1, "description": "..."},
  {"id": 2, "description": "..."}
]
```
\end{Verbatim}
\end{promptbox}

For \methodname{}-Design, the question additionally incorporates the Research Method as the proposed approach and instructs the reader to design a verification experiment for it; the reference answer is written from the Experimental Design stage alone, and all rubrics are elicited along the experimental-design dimension only. The experimental-design task prompt is as follows:

\begin{promptbox}
\begin{Verbatim}[breaklines=true, breakanywhere=false]
# Role
You are an experienced researcher with expertise in designing experimental validation tasks.

# Task
Based on the provided [Research Field], [Research Background], and [Original Solution], construct a clear experimental design task as a single coherent paragraph. The task should:
- Describe the research context, the current state of existing approaches, and key challenges
- Present the Original Solution as an existing method innovation that has been proposed to address the problem
- Clearly instruct the reader to design a reasonable verification experiment to validate the effectiveness of this method
- The task type is: based on existing research question, research background, and method innovation, design a reasonable verification experiment

The Original Solution should be naturally integrated into the task description as the proposed method that needs to be experimentally validated.

# Abstraction Constraints (Critical)
To keep the experimental design space open, the task description must NOT disclose or hint at any experimental validation details:
- Do NOT mention or suggest any specific baselines, comparison methods, datasets, benchmarks, evaluation metrics, ablation settings, control variables, or expected outcomes
- Describe the proposed method only at the level of its core idea, motivation, and technical mechanism; avoid implementation-level specifics that would dictate how the experiment must be conducted
- The reader should be unable to predict the concrete experimental setup (which baselines, which datasets, which metrics, which ablations) from the task description alone

# Output Format
Wrap the final task description inside <output> </output> tags. Rules for the content inside the tags:
- Write as a single paragraph of fluent, flowing prose
- Do NOT use bullet points, numbered lists, or section headers of any kind
- Include NO introductory phrases (e.g., "Here is the task:", "Based on the background:")
- Include NO conversational filler or meta-commentary (e.g., "I have written...", "This statement covers...")
- Output only the task description itself, nothing else

# Input
[Research Field]:
{research_field}

[Research Background]:
{research_background}

[Original Solution]:
{raw_solution}
\end{Verbatim}
\end{promptbox}

The experiment-only reference-answer prompt is as follows:

\begin{promptbox}
\begin{Verbatim}[breaklines=true, breakanywhere=false]
# Role
You are an experienced researcher tasked with writing a reference solution (experimental design) for an experimental design task. Your writing must be rigorous, academic in style, logically structured, and concise.

# Task
Based on the provided [Experimental Design Task] and [Original Experiment], write a reference solution that describes the experimental design. The description must adhere to the following principles:

1. **Faithfulness**: Only describe content that is directly supported by the provided material. Do NOT add any fabricated details, speculative claims, or information not present in the inputs.
2. **No concrete experimental data**: Exclude all specific numerical results, statistics, or quantitative outcomes. Focus on the rationale and design logic of the experiments, not the raw numbers.
3. **Content focus**: Emphasize the motivation behind the experimental design (e.g., choice of baselines, datasets, evaluation metrics, variable controls), the experimental setup, and the reasoning for the approach.
4. **Academic style**: Use formal, precise, and concise academic language. Write in a structured but paragraph-based flow - avoid bullet points, numbered lists, or section headers. Each sentence should convey a clear technical point.

# Output Format
Wrap the final reference solution inside <output> </output> tags. The content within the tags should be the solution only, without introductory phrases, conversational filler, or meta-commentary.

# Input
[Experimental Design Task]:
{research_problem}

[Original Experiment]:
{raw_experiment}
\end{Verbatim}
\end{promptbox}

The experiment-only rubric prompt for \methodname{}-Design restricts the dimension to experimental design:

\begin{promptbox}
\begin{Verbatim}[breaklines=true, breakanywhere=false]
# Role
You are an expert in scientific methodology and solution evaluation. Your task is to analyze the given experimental design task and reference solution, and generate comprehensive evaluation rubrics to assess the quality of any proposed solution in a binary (True/False) manner.

# Experimental Design Task
{research_problem}

# Reference Solution
{reference_solution}

# Rubric Design Guidelines
Each rubric must be:
- **True/False binary**: objectively assessable without subjective interpretation.
- **Atomic**: evaluate ONE and only ONE independent concept per rubric.
- **Objective**: avoid subjective adjectives (good, detailed, clear, reasonable); use deterministic phrasing (e.g., "Contains formula X", "Mentions concept Y").
- **Negation-style positively phrased**: e.g., "No logical conflicts are present" rather than "Logical conflicts occurred".
- **Not about final results**: do NOT include rubrics that require executing experiments to verify.
- **Not about trivial details**: avoid minor/overly specific details of the reference solution, as multiple valid approaches may exist.
- **Non-inferable from the task alone (CRITICAL)**: each rubric must evaluate a specific design decision grounded in the [Reference Solution] that is NOT stated, implied, or predictable from the [Experimental Design Task] alone. Before finalizing each rubric, perform this self-check: "Could a reader who has only read the task description (never the reference solution) already know that a correct answer must satisfy this rubric?" If yes, DISCARD the rubric and design a more discriminating one.
- **Not generic methodology**: forbid boilerplate rubrics that any competent experimental design would trivially satisfy (e.g., "The solution includes baseline comparisons", "The solution uses appropriate evaluation metrics", "The solution controls for confounding variables"). Every rubric must name a concrete, discriminating design element (a specific baseline category, a specific dataset property, a specific ablation, a specific control) whose necessity cannot be guessed without knowing the reference solution.

Construct rubrics focusing solely on the **Experimental Design** dimension:

Assess whether the solution proposes a sound and well-justified experimental setup, including appropriate baselines, datasets, evaluation metrics, variable controls, handling of confounds, and reproducibility considerations.

Prioritize rubrics that target the **core experimental design choices and key validation steps** - discard trivial or peripheral aspects. Fewer high-impact rubrics are better than many shallow ones.

# Output Requirements
Generate exactly 10 rubrics. Avoid overly specific numerical settings (e.g., learning rates, batch sizes, epochs). Provide strictly valid JSON.
Wrap the output JSON array in ```json ``` code blocks, formatted as:
```json
[
  {{"id": 1, "description": "..."}},
  {{"id": 2, "description": "..."}}
]
```
\end{Verbatim}
\end{promptbox}

For both benchmarks, the merge-and-rank prompt is identical to the one in Sec.~\ref{sec:appendix-qa}, except that for \methodname{}-Design the task header is relabeled as ``Experimental Design Task''.

\end{document}